\documentclass[10pt,journal,compsoc]{IEEEtran}

\usepackage{amsmath}

\DeclareMathOperator*{\minimize}{minimize}
\DeclareMathOperator*{\argmax}{argmax}
\DeclareMathOperator*{\argmin}{argmin}
\def\bb{\ensuremath\boldsymbol}

\usepackage{graphicx}
\usepackage{amssymb}
\usepackage{blindtext}
\usepackage{booktabs}
\usepackage{bbm}

\long\def\ignorethis#1{}
\usepackage[usenames,dvipsnames,svgnames,x11names]{xcolor}

\definecolor{gray}{rgb}{0.6,0.6,0.6}
\definecolor{red}{rgb}{1,0,0}
\definecolor{green}{rgb}{0,1,0}
\definecolor{blue}{rgb}{0,0,1}
\definecolor{dark-green}{rgb}{0,0.4,0}
\definecolor{orange}{rgb}{1,0.55,0}
\definecolor{white}{rgb}{1,1,1}
\definecolor{black}{rgb}{0,0,0}
\definecolor{dark-brown}{rgb}{0.2,0.1,0}
\definecolor{light-blue}{rgb}{0.4,0.6,0.99}
\definecolor{dark-red}{rgb}{0.6,0,0}
\definecolor{light-red}{rgb}{1,0.2,0.6}
\definecolor{pink}{rgb}{1,0.2,0.6}
\definecolor{dark-pink}{rgb}{0.6,0,0.3}
\definecolor{darkblue}{rgb}{0.0, 0.4, 0.75}

\newcommand{\whitetext}[1]{{\color{white}#1}\normalfont}

\usepackage{amssymb}
\usepackage{pifont}
\newcommand{\cmark}{\ding{51}}%
\usepackage{wrapfig,lipsum,booktabs}

\usepackage{algorithm}
\usepackage{listings}
\usepackage{etoolbox}
\makeatletter
\AfterEndEnvironment{algorithm}{\let\@algcomment\relax}
\AtEndEnvironment{algorithm}{\kern2pt\hrule\relax\vskip3pt\@algcomment}
\let\@algcomment\relax
\newcommand\algcomment[1]{\def\@algcomment{\footnotesize#1}}
\renewcommand\fs@ruled{\def\@fs@cfont{\bfseries}\let\@fs@capt\floatc@ruled
  \def\@fs@pre{\hrule height.8pt depth0pt \kern2pt}%
  \def\@fs@post{}%
  \def\@fs@mid{\kern2pt\hrule\kern2pt}%
  \let\@fs@iftopcapt\iftrue}
\makeatother

\usepackage[final]{changes}
\definechangesauthor[color=NavyBlue]{PAN}

\ifCLASSOPTIONcompsoc
  \usepackage[nocompress]{cite}
\else
  \usepackage{cite}
\fi
\ifCLASSINFOpdf
\else
\fi
\begin{document}
%
\title{Learning Canonical Embeddings for Unsupervised Shape Correspondence with Locally Linear Transformations}
%
%
%
%

\author{Pan~He,
        Patrick~Emami,
        Sanjay~Ranka, \IEEEmembership{Fellow,~IEEE}, 
        Anand~Rangarajan
\IEEEcompsocitemizethanks{\IEEEcompsocthanksitem P. He, S. Ranka, and A. Rangarajan are with the Department of Computer and Information Science
and Engineering, University of Florida, Gainesville,
FL 32611, USA. P. Emami is with the National Renewable Energy Lab, Golden, CO 80401, USA.\protect\\
\IEEEcompsocthanksitem  E-mail: pan.he@ufl.edu; patrickemami@gmail.com; ranka@cise.ufl.edu; anand@cise.ufl.edu \protect\\


}

}

\IEEEtitleabstractindextext{%
\begin{abstract}
We present a new approach to unsupervised shape correspondence learning between pairs of point clouds. We make the first attempt to adapt the classical locally linear embedding algorithm (LLE)---originally designed for nonlinear dimensionality reduction---for shape correspondence. The key idea is to find dense correspondences between shapes by first obtaining high-dimensional neighborhood-preserving embeddings of low-dimensional point clouds and subsequently aligning the source and target embeddings using locally linear transformations. We demonstrate that learning the embedding using a new LLE-inspired point cloud reconstruction objective results in accurate shape correspondences. More specifically, the approach comprises an end-to-end learnable framework of extracting high-dimensional neighborhood-preserving embeddings, estimating locally linear transformations in the embedding space, and reconstructing shapes via divergence measure-based alignment of probabilistic density functions built over reconstructed and target shapes. Our approach enforces
embeddings of shapes in correspondence to lie in the same universal/canonical embedding space, which eventually helps regularize the learning process and leads to a simple nearest neighbors approach between shape embeddings for finding reliable correspondences. Comprehensive experiments show that the new method makes noticeable improvements over state-of-the-art approaches on standard shape correspondence benchmark datasets covering both human and nonhuman shapes.
\end{abstract}

}

\maketitle

\IEEEdisplaynontitleabstractindextext

%
\IEEEpeerreviewmaketitle

\IEEEraisesectionheading{\section{Introduction}\label{sec:introduction}}

\IEEEPARstart{T}he shape correspondence learning problem is fundamental to geometry processing and computer vision and has been used as a key component in many downstream applications such as deformation modeling \cite{sumner2004deformation}, texture mapping \cite{yin20213dstylenet}, and medical imaging  \cite{shen2001adaptive}, to name a few. 

Dense correspondences between a pair of shapes can be established by measuring the similarities of extracted feature descriptors. Traditional approaches have identified a set of geometric feature descriptors, including extrinsic and intrinsic descriptors \cite{johnson1999using,belongie2006matching,rusu2009fast,rustamov2007laplace,sun2009concise,aubry2011wave}. However, these handcrafted descriptors often lead to inaccurate and time-consuming solutions. More recently, we have seen the emergence of \textit{data-driven} approaches built upon modern machine learning techniques that learn the optimal features directly from massive shape pair datasets \cite{litman2013learning,rodola2014dense,boscaini2016learning,monti2017geometric,marin2020correspondence,cosmo2020average}. However, the major dissatisfaction here is a need for supervised learning, which relies on a sufficient number of labeled training pairs of high-quality ground truth correspondences, which are known to be scarce and difficult to obtain. By contrast, the unsupervised approaches \cite{groueix20183d,roufosse2019unsupervised,zeng2021corrnet3d,lang2021dpc,chen2021unsupervised} seek to remove the dependency on ground truth correspondence by employing autoencoder-inspired architectures, where they construct the deformation between a pair of shapes and leverage point reconstruction to learn suitable features for measuring the similarities between shapes. However, most of them suffer from the nontrivial optimization of the deformation and reconstruction, thus often requiring additional regularization or constraints, e.g., cycle consistency \cite{ginzburg2020cyclic} and local smoothness \cite{lang2021dpc}, and usually achieving limited generalization performance.
\begin{figure*}[t]
    \centering
    \includegraphics[width=\textwidth]{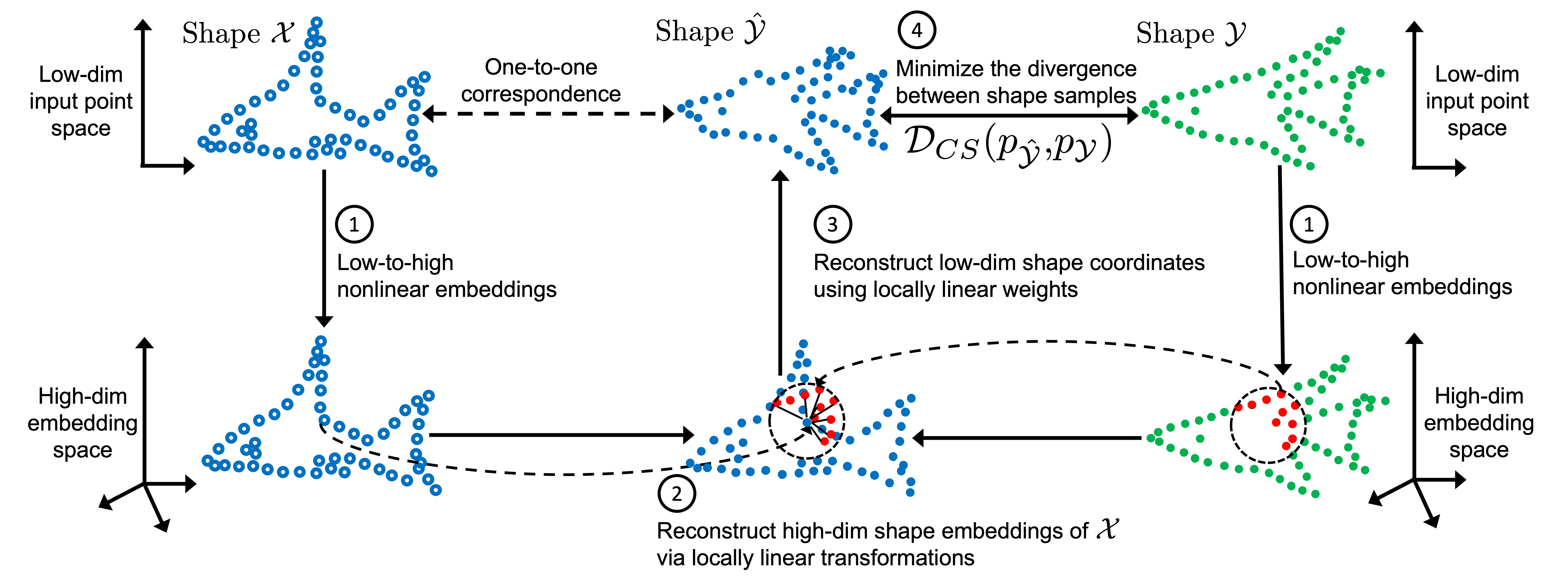}
    \caption{Given the source and target shapes  ($\mathcal{X}$ and $\mathcal{Y}$) as input: (1) We first project the low-dimensional point cloud coordinates into the high-dimensional embedding space; (2) We align feature embeddings of $\mathcal{X}$ and $\mathcal{Y}$ by computing the optimal locally linear transformation that can best cross-reconstruct each feature embedding of $\mathcal{X}$ using its top-K nearest neighbors (red color) in the embedding space of $\mathcal{Y}$; (3) We associate the embeddings of these neighbors to their original point cloud coordinates and further leverage the reconstruction weights of the optimal transformation to reconstruct a shape $\hat{\mathcal{Y}}$ sharing the same point indices to $\mathcal{X}$ (both in blue color); and (4) The embedding network can be optimized by minimizing the divergence between $\hat{\mathcal{Y}}$ and $\mathcal{Y}$.}
    \label{fig:conceptual}
    \vspace{2mm}
\end{figure*}


In this paper, we present a new unsupervised learning framework for shape correspondence between pairs of point clouds. Inspired by the classical locally linear embedding (LLE) algorithm \cite{roweis2000nonlinear} initially used for nonlinear dimensionality reduction, we make the first attempt to adapt this concept for shape correspondence. LLE succeeds in exploiting the local Euclidean geometry of manifolds to approximately preserve the local Euclidean geometry within neighborhoods. In a similar manner, because a point cloud shape is a sampled version of a smooth manifold, it is desirable to learn shape embeddings capable of capturing the underlying structure of the manifold. Our method achieves this by learning high-dimensional neighborhood-preserving embeddings of low-dimensional shapes such that nearby and corresponding points fall close to each other in both the high-dimensional embedding space \textit{and} in the low-dimensional input space. 

Another point of departure is that in recent approaches like functional maps, non-rigid deformations between shapes are expected to become \textit{linear} transformations once shapes are projected into a higher-dimensional embedding space. Essentially, point-to-point correspondences between shapes are generalized as a \textit{linear map} between the corresponding function spaces \cite{ovsjanikov2012functional}. 
It is worth emphasizing that our approach is fundamentally different from the functional map-based approaches \cite{ovsjanikov2012functional,marin2020correspondence}. These approaches interpret the basis as the embedding for each shape and represent the mapping between a pair of shapes as a change of basis matrix, by applying a \textit{global linear transformation} to every shape point. By contrast, our approach treats maps between shapes as \textit{locally linear transformations} between embeddings, where each shape point has its own linear transformation computed from local neighboring regions. The locally linear transformations succeed in identifying the underlying structure of the shape manifold by enforcing embeddings of shapes in correspondence to lie in the same universal/canonical embedding space, which eventually helps regularize the learning process and leads to a simple nearest
neighbors approach for finding reliable correspondences.

We achieve our goals through the following steps, all driven by the idea of marrying LLE and the construction of high-dimensional nonlinear embeddings of point cloud shapes (Figure \ref{fig:conceptual}). Assume we have a nonlinear embedding structure taking a low-dimensional point cloud and returning a high-dimensional embedding, whose weights we seek to learn. Given two shapes whose correspondence we seek, in the first step, we attempt to cross-reconstruct each source point (in the embedding space) from its nearest neighbors in the target point cloud embedding. This step mirrors the first stage of LLE. Next, we take the obtained reconstruction weights and cross-reconstruct a point cloud shape (in the original low-dimensional space) from the same set of nearest neighbors in the target point cloud (again in the original low-dimensional space). We thereby obtain a reconstructed point cloud in one-to-one correspondence with the source point cloud but based on the nearest neighbors in the target point cloud. We then minimize a suitable divergence measure between the cross-reconstructed and target point clouds with respect to the unknown weights of the nonlinear embedding. Minimization of the cross-reconstruction error minimizes the distance between the original and reconstructed point clouds, and minimization of the divergence measure brings the cross-reconstruction and target point clouds into register. In this way, we build an unsupervised shape correspondence engine capable of end-to-end learning of  nonlinear universal embeddings of shape point clouds.
\subsection{Contributions}
In summary, our contributions are: 
\begin{itemize}
    \item A new perspective on finding dense correspondences between shapes as \textit{locally linear} transformations in a high-dimensional embedding space, as a superior way to regularize the embeddings of shapes in correspondence via forcing them to lie in the same canonical embedding space 
    \item An unsupervised shape correspondence learning framework for extracting nonlinear shape embeddings that preserve distances within local neighborhoods, estimating locally linear transformations in the embedding space, and reconstructing shapes via the alignment of probability density functions (PDFs) built over reconstructed and target shapes
    \item A divergence measure for bringing the cross-reconstructed and target shape PDFs into register, which shows improved performance over popular Chamfer distance (CD) and Earth mover's distance (EMD) measures
    \item A significant improvement compared to existing state-of-the-art methods on standard benchmarks covering both human and nonhuman shapes. 
\end{itemize}
 
Comprehensive experiments show that the new method makes substantial improvements while showing strong model generalization across datasets with efficient training and inference. More importantly, the proposed idea could be useful for matching problems in other modalities such as images and meshes and matching problems in cross-modality such as images
to point clouds, and it is a promising approach for other tasks requiring the application of manifold learning concepts.

\section{Related Work}

 \added[id=PAN,comment={}]{\noindent \textbf{Shape Correspondence and Matching.} Early efforts at representing correspondence (before the deep learning era) used the inexact weighted graph matching formulation with a permutation matrix and outlier for correspondence representation \cite{carcassoni2003spectral,luo2003unified,knossow2009inexact}, which is further followed by simultaneous pose and correspondence estimation. Simpler (but not necessarily better) methods such as Iterative closest point (ICP) \cite{besl1992method} and Chamfer matching (CM) \cite{thayananthan2003shape}  started seeing deployment, followed by the emergence of Earth mover’s
distance (EMD) \cite{levina2001earth} and transportation-based distance measures. Simultaneously, soft correspondences via
softassign \cite{rangarajan1997convergence,rangarajan1997softassign,chui2003new} for both linear assignment and quadratic assignment alternatively estimated  the transformations and  updated the explicit point-to-point correspondence. Coherent Point Drift (CPD) \cite{myronenko2010point} is similar to Robust Point Matching (RPM) \cite{chui2003new} and used Gaussian radial basis functions (GRBF) instead of thin-plate splines (TPS) for non-rigid deformations. RPM L$_2$E \cite{ma2013robust,ma2015robust} leveraged the L$_2$E estimator for estimating transformations bootstrapped from the shape context \cite{belongie2002shape}. Later, point cloud density estimation approaches \cite{fitzgibbon2003robust,tsin2004correlation,roy2007deformable,jian2010robust,hasanbelliu2011robust} have appeared coupled with distances between density functions optimized
w.r.t. the unknown spatial transformation without establishing the explicit point correspondence. Representative approaches include KC \cite{tsin2004correlation}, GMMReg \cite{roy2007deformable,jian2010robust}, and CS \cite{hasanbelliu2011robust}, to name a few.}

The functional map was introduced in the pioneering work of Ovsjanikov at al. \cite{ovsjanikov2012functional} for solving non-rigid shape matching by avoiding the direct estimation of point-to-point correspondence and instead modeling linear transformations between the functional spaces of shapes, which is followed by subsequent extensions such as \cite{kovnatsky2015functional,kovnatsky2016madmm,litany2017deep,rodola2017partial,roufosse2019unsupervised,pai2021fast}. Recently, Diff-FMaps \cite{marin2020correspondence}  interprets the eigendecomposition based on the Laplace Beltrami Operator (LBO)  as higher-dimensional embeddings of shapes. More importantly, it demonstrated that learning a canonical embedding is a nontrivial problem, and splitting the correspondence learning into two parts (\textit{invariant embedding + linear transformation}) is beneficial in regularizing the embedding learning in challenging settings. Meanwhile, we have also witnessed  progress in the \textit{unsupervised} functional map approaches \cite{halimi2019unsupervised,roufosse2019unsupervised,ginzburg2020cyclic} that consider structural penalties on the inferred maps, e.g.,  bijectivity or orthogonality. 

 The spatial approach is another direction in related work. 3D-CODED \cite{groueix20183d} and Elementary \cite{deprelle2019learning} matched the deformable shapes by jointly encoding shapes and correspondences via deforming templates. CorrNet3D \cite{zeng2021corrnet3d}  exploited DGCNN \cite{wang2019dynamic} to project shapes into a high-dimensional feature space. They enforced unsupervised feature learning by constructing a symmetric deformer for point cloud reconstruction.  \added[id=PAN,comment={}]{Trappolini et al. \cite{trappolini2021shape} proposed a transformer-based framework to efficiently estimate the transformation between point cloud shapes.}
 \added[id=PAN,comment={}]{DPC \cite{lang2021dpc} demonstrated a self- and cross-reconstruction framework to learn the latent affinity via a simplified point reconstruction, which is completely different from existing encoder-decoder frameworks \cite{groueix20183d,zeng2021corrnet3d} regressing ordered point clouds to determine matching points.} DPC normalized the similarity of each feature embedding's K nearest neighbors via softmax and  cross-reconstructed a shape using the corresponding input points and similarity scores. Additional mapping loss and self-reconstruction have to be included to impose smooth constraints, otherwise, the performance of DPC drops significantly without them (see their ablation study on design choices). 
 \added[id=PAN,comment={}]{In this paper, we focus on learning LLEs capable of capturing the underlying structure of the shape manifold, with the main goal to find a proper design of the embedding via leveraging local neighborhood relations.} Our method simultaneously learns the embedding and optimal locally linear transformation using the LLE-inspired first-stage method and point reconstruction. Consequently, the cross-reconstruction of the source shape using target points gets an implicit regularization via the close relationship to the reconstruction of the source's high-dimensional embedding counterparts. We conjecture that this is exactly what is missing in DPC and hence explains our superior performance using just cross-reconstruction.

\noindent \textbf{Shape Descriptor and Feature Learning.} 
The shape analysis community has actively investigated extracting descriptors and feature maps from shapes to capture geometric properties around the neighborhood of points of interest.
A detailed discussion of classic \textit{hand-crafted} descriptors can be found in \cite{bustos2005feature,tangelder2008survey}. 

\added[id=PAN,comment={}]{Early attempts focused on invariance under a global spatial transformation, e.g., a rigid motion, as shown in shape context \cite{belongie2002shape}, spin image \cite{johnson1999using}, and multiscale local feature \cite{pauly2003multi}. The community has then extended to the nonrigid case by considering geodesic distances
and conformal factors \cite{hamza2003geodesic,lipman2009mobius}. Later, the \textit{diffusion geometry} \cite{berard1994embedding} established invariant metrics based on eigenvalues and eigenvectors of the Laplace Beltrami operator obtained from shapes, showing significantly more robustness compared to the geodesic counterparts \cite{memoli2009spectral,bronstein2010gromov}. Follow-ups include global point signature (GPS) \cite{rustamov2007laplace}, heat kernel signature (HKS) \cite{sun2009concise},
 and wave kernel signature (WKS) \cite{aubry2011wave}.}

The \textit{data-driven} feature descriptors have shown their advantages over the \textit{hand-crafted} features in robustness and efficiency.  
The bag-of-features (BoF) descriptor extracted the frequency histograms of geometric words from shapes \cite{bronstein2011shape}. 
In \cite{castellani2008sparse}, a robust
and invariant point signature is learned from the contextual 3D neighborhood information of salient points for shape matching. Recently, the community has started to extract deep features from shapes in a \textit{data-driven} fashion. For example, GCNN captured invariant shape features from triangular meshes \cite{masci2015geodesic}. PointNet \cite{qi2017pointnet} showed that the learned features could be used
to compute shape correspondences. FMNet \cite{litany2017deep} leveraged a Siamese residual network \cite{he2016deep} for descriptor learning. SplineCNN \cite{fey2018splinecnn} introduced a novel convolution operator
based on B-splines to filter the geometric input efficiently. DeepGFM \cite{donati2020deep} used KPConv \cite{thomas2019kpconv}, a classic point cloud convolutional filter, to extract robust shape features. 

Our approach belongs to the  \textit{data-driven} approaches. To the best of our knowledge, our approach is the first attempt to adapt the classic LLE algorithm for unsupervised shape correspondence learning into an end-to-end learnable framework.  The correspondence is determined via nearest neighbor searching once we learn such discriminative feature representations.

\section{Locally Linear Embedding}

Before describing the proposed approach, we provide a background on the classic LLE framework \cite{roweis2000nonlinear}.
Given input point features, LLE has three steps: first, it identifies $K$ nearest neighbors for each point; second, it  uses least-squares to compute the weights for  reconstructing each point from its nearest neighbors; and third, it reuses the same set of weights and computes the embedding of each point in a low-dimensional space. Specifically, let $\mathrm{X} = \{\bb{x}_i \in \mathbb{R}^D | i = 1, \dots, N \}$ denote the input point set. LLE builds a K nearest neighbors (KNN) graph over $\mathrm{X}$ by measuring the  pairwise Euclidean distance and removing self-loops from the graph. Denoting $\bb{x}_{il} \in \mathbb{R}^D$ as the $l$-th nearest neighbor of point $\bb{x}_i$, we obtain the LLE reconstruction weights by solving
\begin{equation}
\begin{split}\label{eq:lsq}
    &\minimize_{\boldsymbol{W}} \quad E(\boldsymbol{W}) = \sum_{i=1}^N \Big\Vert \bb{x}_i - \sum_{l=1}^K  w_{il} \bb{x}_{il} \Big\Vert_2^2 \\
    & \textrm{subject to} \quad  \sum_{l=1}^K w_{il} = 1, \forall i \in \{1, \dots, N\},
\end{split}    
\end{equation}
where $\mathbb{R}^{N \times K} \ni \boldsymbol{W} := [\bb{w}_1, \dots, \bb{w}_N]^T$ denotes the reconstruction weights, and $\mathbb{R}^K \ni \bb{w}_i := [w_{i1}, \dots, w_{iK}]^T$ denotes the weights associated to the KNN neighbors $\{\bb{x}_{il}\}_{l=1}^K$ for reconstructing point $\bb{x}_i$. The sum-to-one weight constraint $\sum_{l=1}^K w_{il} = 1$ leads to the specific properties---each point  $x_i$ is invariant to rotations, translations, and rescalings of itself and its nearest neighbors \cite{roweis2000nonlinear}. The optimal $\boldsymbol{W}$ can be found by solving a constrained least squares problem as detailed in Appendix \ref{sec:lsq}. 
Given the optimal $\boldsymbol{W}$, LLE further finds the lower-dimensional embeddings of input points denoted as $\mathrm{Y} = \{\bb{y}_i \in \mathbb{R}^{D_2} | i = 1, \dots, N \}, D_2 \ll D$, by solving 
a sparse eigenvalue problem\cite{roweis2000nonlinear}.



\section{LTENet}

\begin{figure*}
    \centering
    \includegraphics[width=\textwidth]{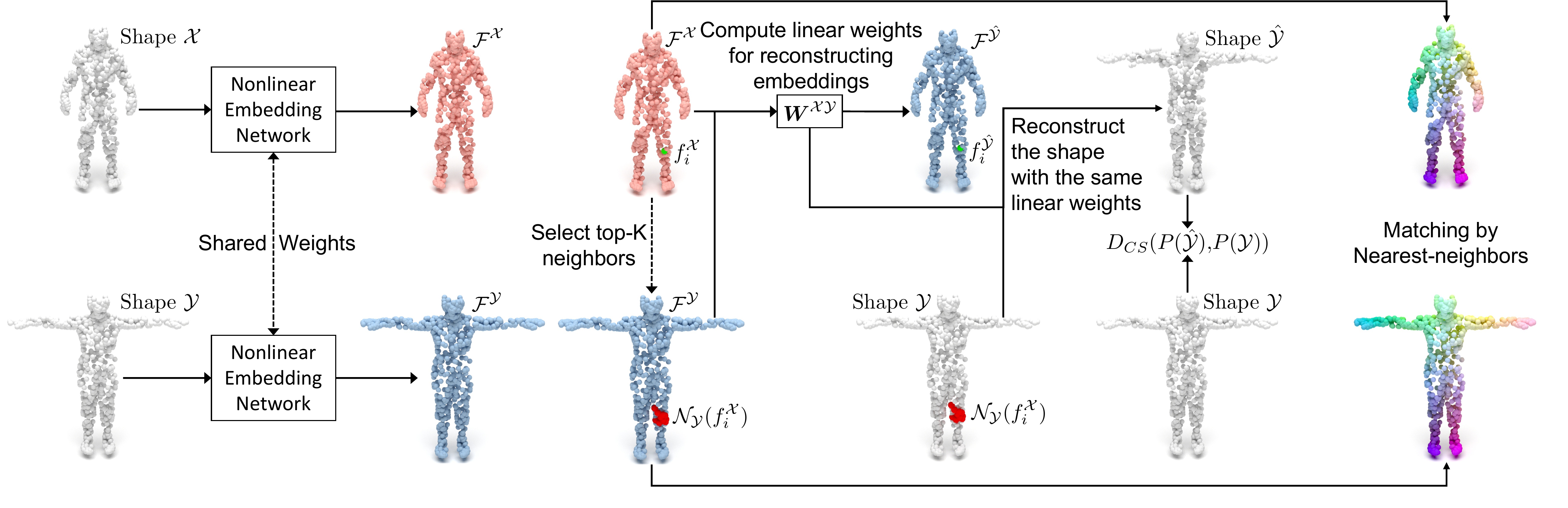}
    \caption{The pipeline overview of LTENet: (1) extract nonlinear shape embeddings $\mathcal{F}^\mathcal{X}$ and $\mathcal{F}^\mathcal{Y}$, given $\mathcal{X}$ and $\mathcal{Y}$; (2) select top-K neighbors for each feature embedding $f^\mathcal{X}_i$ based on the cosine similarity between $\mathcal{F}^\mathcal{X}$ and $\mathcal{F}^\mathcal{Y}$; (3) estimate the locally linear transformations following Equations \ref{eq:cross_reverse_weight} and \ref{eq:phi_function} to best reconstruct $\mathcal{F}^\mathcal{X}$ using $\mathcal{F}^\mathcal{Y}$ and denoted as $\mathcal{F}^{\hat{\mathcal{Y}}}$; (4) reconstruct a shape $\hat{\mathcal{Y}}$ following Equation \ref{eq:cross_reconstruction}; (5) learn the embedding network $\mathcal{F}$ via minimizing the divergence $\mathcal{D}_{CS}  (P(\hat{\mathcal{Y}}), P(\mathcal{Y} ))$; and (6) determine the correspondence using nearest neighbors between embeddings.}
    \label{fig:main}
\end{figure*}

Having briefly summarized LLE, we introduce our novel approach to unsupervised shape correspondence learning called \textbf{LTENet}---\textbf{L}ocally Linear \textbf{T}ransformation based \textbf{E}mbedding \textbf{Net}works. To the best of our knowledge, this is the first attempt at introducing an LLE-inspired algorithm that represents maps between pairs of shapes as locally linear transformations while simultaneously deploying an LLE shape reconstruction objective to optimize nonlinear embeddings towards the same universal/canonical space. In this section, we first define the shape correspondence problem. We then introduce the dovetailing of locally linear transformations with novel LLE point cloud
reconstructions for learning both the optimal transformation and embedding. Finally, we describe the divergence measure between the cross-reconstructed and target point clouds.
The  pipeline of LTENet is summarized in Figure~\ref{fig:main}.

\subsection{Problem Definition and Objectives}
Let point clouds  $\mathbb{R}^{N \times 3} \ni \mathcal{X} := [\bb{x}_1, \dots, \bb{x}_N]^T$ and $\mathbb{R}^{N \times 3} \ni \mathcal{Y} := [\bb{y}_1, \dots, \bb{y}_N]^T$ denote the source and target shapes, respectively, where $\bb{x}_i, \bb{y}_j \in \mathbb{R}^3$ and $N$ is the number of points. Our goal is to find a point-to-point correspondence or map defined as $T_{\mathcal{XY}}: \mathcal{X} \rightarrow \mathcal{Y}$ such that every point $\bb{x}_i$ in $\mathcal{X}$ has its corresponding point $\bb{y}_{j^*} := T_\mathcal{XY}(\bb{x}_i)$ in $\mathcal{Y}$, where $1 \leq i, j^* \leq N$. 
Inspired by LLE \cite{roweis2000nonlinear}, the proposed LTENet leads to suitable embeddings for shape correspondence that preserve the local configurations of nearest neighbors. From a high-level perspective, it shares a similar spirit with the LBO operator which relies on preserving distances between nearby points. While the LBO operator is usually constructed over point cloud coordinates, LTENet operates on nonlinear feature embeddings obtained from deep neural networks, which are more robust and efficient to compute. 
It is worth mentioning that the proposed approach directly takes raw point cloud coordinates as the input without containing any point connectivity information. 

\subsection{Optimal Locally Linear Transformations}

Given $\mathcal{X}$ and $\mathcal{Y}$, we extract their  nonlinear feature embeddings $\mathcal{F}^\mathcal{X},\mathcal{F}^\mathcal{Y} \in \mathbb{R}^{N\times D}$, respectively, via a neural network $\mathcal{F}$. This allows us to transition from the point cloud coordinates ($\mathbb{R}^{N \times 3}$) to a higher dimensional embedding
space ($\mathbb{R}^{N\times D})$ where finding shape correspondence is more likely to be successful as demonstrated by the functional map paradigm \cite{ovsjanikov2012functional}.

To align $\mathcal{F}^\mathcal{X}$ and $\mathcal{F}^\mathcal{Y}$, we must find a transformation between them while also jointly optimizing $\mathcal{F}$ to obtain suitable embeddings for estimating shape correspondence.
Recall that Equation~(\ref{eq:lsq}) in LLE is able to approximate the original input points using the LLE reconstruction weights and nearest neighbors. 
Denote $\mathcal{F}^{\hat{\mathcal{Y}}} \in \mathbb{R}^{N\times D} $ as the feature embeddings obtained by applying the transformation to $\mathcal{F}^\mathcal{Y}$ to achieve the alignment, i.e., $\mathcal{F}^\mathcal{X} \approx  \mathcal{F}^{\hat{\mathcal{Y}}}$.
Given all $f^\mathcal{X}_i  \in \mathcal{F}^\mathcal{X}$ and $f^\mathcal{Y}_j  \in \mathcal{F}^\mathcal{Y}$, we implement the transformation by first considering
\begin{equation}
\begin{split}\label{eq:cross_reverse_weight}
    &\minimize_{\boldsymbol{W}^{\mathcal{X}\mathcal{Y}}} \quad E(\boldsymbol{W}^{\mathcal{X}\mathcal{Y}}) = \sum_{i=1}^N \Big\Vert f^\mathcal{X}_i - \sum_{l \in \mathcal{N}_{\mathcal{Y}}(f^\mathcal{X}_i)}  w_{i,l}^{\mathcal{X}\mathcal{Y}} f^\mathcal{Y}_l \Big\Vert_2^2 \\
    & \textrm{subject to} \quad  \sum_{l \in \mathcal{N}_{\mathcal{Y}}(f^\mathcal{X}_i)} w_{i,l}^{\mathcal{X}\mathcal{Y}} = 1, \forall i \in \{1, \dots, N\},
\end{split}    
\end{equation}
where $\mathcal{N}_{\mathcal{Y}}(f^\mathcal{X}_i)$ are all $K$ indices of $f^\mathcal{X}_i$'s nearest neighbors in $\mathcal{F}^\mathcal{Y}$ based on the cosine similarity.  $\mathbb{R}^{N \times K} \ni \boldsymbol{W}^{\mathcal{X}\mathcal{Y}} := [\bb{w}_1^{\mathcal{X}\mathcal{Y}}, \dots, \bb{w}_N^{\mathcal{X}\mathcal{Y}}]^T$ denotes the reconstruction weights, and $\bb{w}_i^{\mathcal{X}\mathcal{Y}} \in \mathbb{R}^K$ stacks all the reconstruction weights $w_{i,*}^{\mathcal{X}\mathcal{Y}}$ associated to $f^\mathcal{X}_i$ and $\mathcal{N}_{\mathcal{Y}}(f^\mathcal{X}_i)$. We then express each transformed  $ f^{\hat{\mathcal{Y}}}_i \in  \mathcal{F}^{\hat{\mathcal{Y}}}$ as
\begin{equation}\label{eq:phi_function}
    f^{\hat{\mathcal{Y}}}_i = \sum_{l \in \mathcal{N}_{\mathcal{Y}}( f^\mathcal{X}_i)}  w_{i,l}^{\mathcal{X}\mathcal{Y}} f^\mathcal{Y}_l \approx f^\mathcal{X}_i.
\end{equation} As observed from Equations (\ref{eq:cross_reverse_weight}) and (\ref{eq:phi_function}),  we make two unique modifications compared to Equation (\ref{eq:lsq}) of the original LLE paper: (1) We directly operate on high-dimensional nonlinear feature embeddings rather than raw input data, and (2) unlike LLE which reconstructs the input data by picking nearest neighbors from itself, we conduct a cross-reconstruction such that the feature embeddings $\mathcal{F}^\mathcal{X}$ of the source shape will select nearest neighbors from $\mathcal{F}^\mathcal{Y}$ in the target shape, which enforces embeddings of shapes in correspondence to lie in the same universal/canonical embedding space.

Intuitively, Equation (\ref{eq:cross_reverse_weight}) represents the spatial transformation between shapes, e.g., a non-rigid transformation, in the input point space as an equivalent locally linear transformation between $\mathcal{F}_\mathcal{X}$ and $\mathcal{F}_\mathcal{Y}$.  An optimal $\boldsymbol{W}^{\mathcal{X}\mathcal{Y}}$ implies that $\mathcal{F}^\mathcal{X}$ has been properly reconstructed by $\mathcal{F}^{\hat{\mathcal{Y}}}$ using  $\mathcal{F}^\mathcal{Y}$. Following LLE \cite{roweis2000nonlinear,ghojogh2020locally}, the optimal $\boldsymbol{W}^{\mathcal{X}\mathcal{Y}}$ can be found by solving a constrained least squares problem:
\begin{align}\label{eq:deep_lsq}
\bb{w}_i^{\mathcal{X}\mathcal{Y}} = \frac{(\bb{G}_i^{\mathcal{X}\mathcal{Y}} + \gamma \bb{I})^{-1} \bb{1}}{\bb{1}^T (\bb{G}_i^{\mathcal{X}\mathcal{Y}}+ \gamma \bb{I})^{-1} \bb{1}},
\end{align}
where $\bb{I}$ is the identity matrix and $ \bb{1} \in \mathbb{R}^{K  \times 1}$ is the matrix  filling all elements with one. $\bb{G}_i^{\mathcal{X}\mathcal{Y}}$ denotes the Gram matrix defined as
\begin{equation}\label{eq:deep_gram}
\mathbb{R}^{K  \times K} \ni \bb{G}_i^{\mathcal{X}\mathcal{Y}} := (f^\mathcal{X}_i \bb{1}^T  - \boldsymbol{\eta}^\mathcal{Y}_i)^T  (f^\mathcal{X}_i  \bb{1}^T  - \boldsymbol{\eta}^\mathcal{Y}_i),
\end{equation}
where $ \boldsymbol{\eta}^\mathcal{Y}_i \in \mathbb{R}^{D \times K}$ stack all feature embeddings of the $K$ neighbors of $f^\mathcal{X}_i$ found in $\mathcal{F}^\mathcal{Y}$.  Adding $\gamma \bb{I}$ to Equation (\ref{eq:deep_lsq}) leads to numerically
stable solutions by avoiding the possible singularity of $\bb{G}_i^{\mathcal{X}\mathcal{Y}}$ \cite{winlaw2011robust,ghojogh2020locally}. This also links our work to Robust LLE with an $\ell_2$ norm-based regularization (see Appendix \ref{sec:lsq} for details).

\subsection{Learning Canonical Embeddings}

The cross-reconstruction weights $\boldsymbol{W}^{\mathcal{X}\mathcal{Y}}$ allow us to represent $\mathcal{F}_\mathcal{X}$ in terms of $\mathcal{F}_\mathcal{Y}$. However, this does not necessarily imply $\boldsymbol{W}^{\mathcal{X}\mathcal{Y}}$ will lead to a better embedding network $\mathcal{F}$ suitable for shape correspondence. To show that, we observe that the closed-form expression of $\boldsymbol{W}^{\mathcal{X}\mathcal{Y}}$ only depends on the Gram matrix $\bb{G}^{\mathcal{X}\mathcal{Y}}$, where each $\bb{G}_i^{\mathcal{X}\mathcal{Y}}$ is constructed based on the feature difference between $f^\mathcal{X}_i$ and $\boldsymbol{\eta}^\mathcal{Y}_i$. Therefore, the optimal $\boldsymbol{W}^{\mathcal{X}\mathcal{Y}}$ essentially relies on $\mathcal{F}^\mathcal{X}$ and $\mathcal{F}^\mathcal{Y}$. $\boldsymbol{W}^{\mathcal{X}\mathcal{Y}}$ is optimal in terms of the reconstruction of  $\mathcal{F}^\mathcal{X}$ using $\mathcal{F}^\mathcal{Y}$ but is not optimal for shape correspondence if no additional optimization step is applied.

\begin{figure*}[t]
    \centering
    \includegraphics[width=0.95 \textwidth]{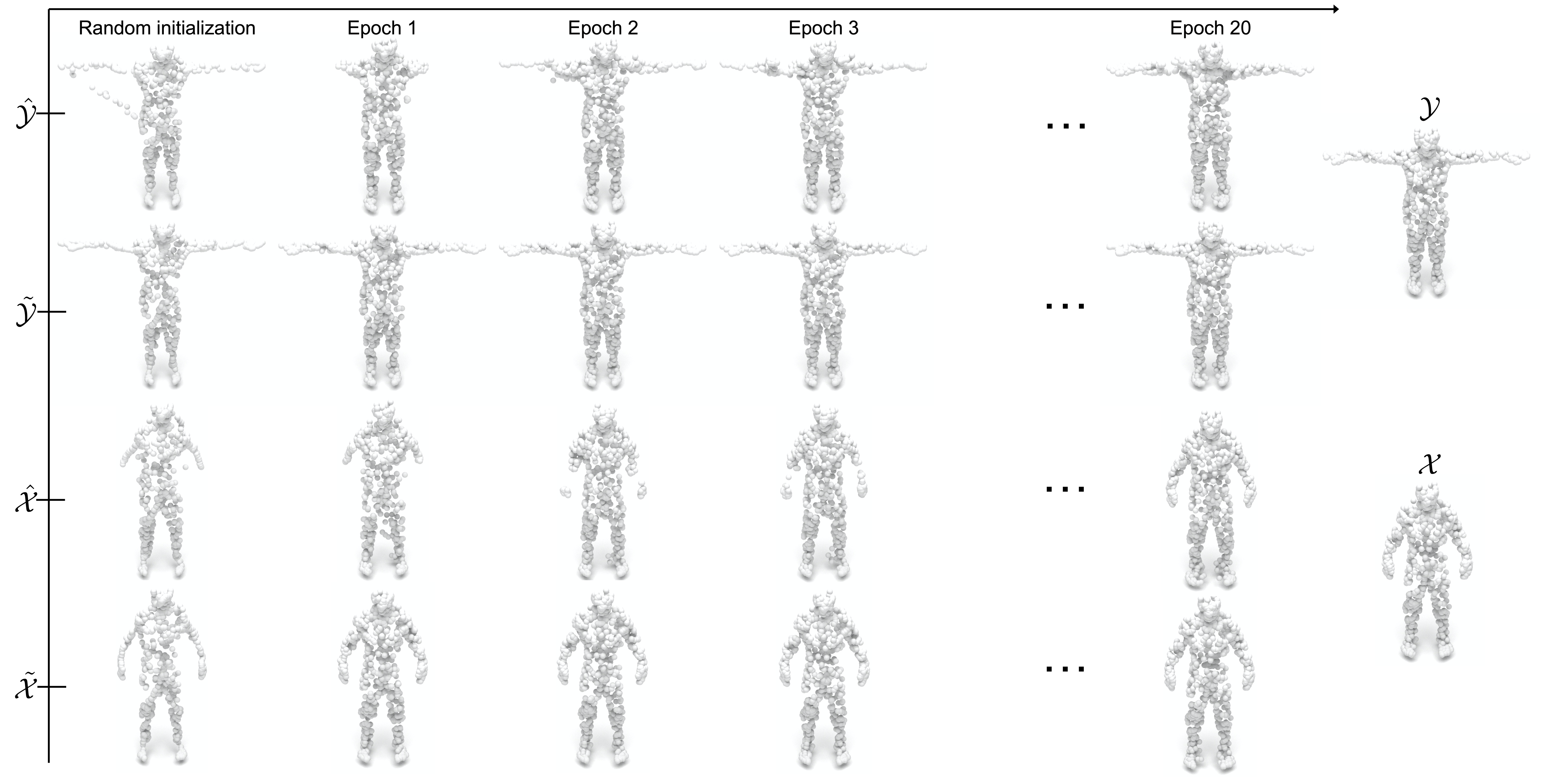}
    \caption{\textbf{Visualization of reconstructed point clouds of the proposed LTENet}. We obtain the reconstructed shapes in cross-reconstruction ($\hat{\mathcal{X}}$ and $\hat{\mathcal{Y}}$) and self-reconstruction ($\tilde{\mathcal{X}}$ and $\tilde{\mathcal{Y}}$) from models at different training epochs. Starting from random initialization on the embedding network, it is clear that the reconstructed shapes are getting closer to the source and target shapes $\mathcal{X}$ and $\mathcal{Y}$ as the training progresses.}
    \label{fig:rec}
\end{figure*}

To properly train the embedding network, we associate the high-dimensional embeddings to the original low-dimensional point cloud coordinates. Specifically, observing $f^{\hat{\mathcal{Y}}}_i = \sum_{l \in \mathcal{N}_{\mathcal{Y}}( f^\mathcal{X}_i)}  w_{i,l}^{\mathcal{X}\mathcal{Y}} f^\mathcal{Y}_l \approx f^\mathcal{X}_i$, we can interpret each feature embedding $f^\mathcal{X}_i$ or $f^{\hat{\mathcal{Y}}}_i$  as a linear combination of a few basis elements $\{f^\mathcal{Y}_l | l \in \mathcal{N}_{\mathcal{Y}}( f^\mathcal{X}_i) \}$ with the associated coefficients $\{w_{i,l}^{\mathcal{X}\mathcal{Y}} | l \in \mathcal{N}_{\mathcal{Y}}( f^\mathcal{X}_i)\}$. Using the same coefficients, we could reconstruct the low-dimensional point $\hat{\bb{y}}_i \in \mathbb{R}^3$ for each $f^{\hat{\mathcal{Y}}}_i$, which gives
\begin{equation}\label{eq:cross_reconstruction}
    \hat{\bb{y}}_i = \sum_{l \in \mathcal{N}_{\mathcal{Y}}( f^\mathcal{X}_i)} w_{i,l}^{\mathcal{X}\mathcal{Y}} \bb{y}_l
\end{equation}
where $\bb{y}_l \in \mathcal{Y}$ is the point associated to $f^\mathcal{Y}_l$. We then obtain $\mathbb{R}^{N \times 3} \ni  \hat{\mathcal{Y}} := [\hat{\bb{y}}_1, \dots, \hat{\bb{y}}_n]^T$, interpreted as the linearly reconstructed shape for $\mathcal{F}^{\hat{\mathcal{Y}}}$ in the basis elements (point coordinates) of $\mathcal{Y}$. The indices of $\hat{\mathcal{Y}}$ are in \textit{exact} one-to-one correspondence with the indices of $\mathcal{X}$. Also, $\hat{\bb{y}}_i$ can be understood as a \textit{soft} correspondence of $\bb{x}_i$ because the indices of $\{\bb{y}_l | l \in \mathcal{N}_{\mathcal{Y}}( f^\mathcal{X}_i)\}$ are the same indices of the top-K nearest neighbors of $f^\mathcal{X}_i$. Because these neighbors are selected from $\mathcal{F}^\mathcal{Y}$ with the highest similarity to $f^\mathcal{X}_i$, the point $\bb{y}_l$ associated to each neighbor could be a candidate matching point of $\bb{x}_i$. 
If each point $\bb{x}_i$  finds its approximate matching point $\hat{\bb{y}}_i$, we would expect $\hat{\mathcal{Y}}$ to be similar to $\mathcal{Y}$ as the training progresses (Figure~\ref{fig:rec}). To this end, we can train $\mathcal{F}$ by solving
\begin{equation}
    \minimize_{\mathcal{F}} \quad E(\mathcal{F}) = \mathcal{D} (\hat{\mathcal{Y}},  \mathcal{Y})
\end{equation} 
where $\mathcal{D} ( \cdot , \cdot)$ defines a dissimilarity measure. Because $\hat{\mathcal{Y}}$ and $\mathcal{X}$ share the same point indices while being different from those of $\mathcal{Y}$, we do not have an obvious one-to-one correspondence between $\hat{\mathcal{Y}}$ and $\mathcal{Y}$, e.g., $\hat{\bb{y}}_1$ probably does not correspond to $\bb{y}_1$, and so on.

\noindent \textbf{LLE vs. LTENet}. LLE finds the embedded vector for each input point by solving an expensive eigenvalue problem (a projection from the high-dimensional input space to the low-dimensional embedding space). In contrast, our approach approximates \textit{source} points via shape reconstruction using linear combinations of nearest neighbor \textit{target} point coordinates $\mathcal{Y}$  and the (closed-form) weights $\boldsymbol{W}^{\mathcal{X}\mathcal{Y}}$ obtained from first stage LLE. Unlike LLE, which is focused on identifying suitable low-dimensional embeddings for high-dimensional input data, LTENet establishes dense  correspondences between shapes by forcing a pair of shapes to lie on the same manifold. This is achieved by our fully differentiable LTENet framework, which pushes their embeddings towards a locally linearly invariant space via maximizing the similarity between a shape and its reconstructed counterpart. LTENet demonstrates a principled approach by adopting the central concept of classic LLE for shape correspondence. Next, we introduce a suitable distance measure $\mathcal{D} ( \cdot , \cdot)$ for end-to-end training.

\subsection{Implicit Correspondence Learning via the Alignment of PDFs}

Most unsupervised approaches  \cite{lang2021dpc} adopt popular CD and EMD measures to reconstruct point clouds, which are sensitive to outliers or are computationally intensive. Point clouds are quite often nothing but discrete samples of the underlying continuous shapes and surfaces. Therefore, we instead represent point clouds as probability density functions and seek to minimize a divergence between reconstructed and original shapes. Formally, given the shape $\mathcal{X} = [\bb{x}_1, \dots, \bb{x}_N]^T$, we represent an arbitrary point $\bb{x}$ by its kernel (Parzen) density estimation (KDE) of the PDF using an arbitrary kernel function $\mathsf{K}(\cdot)$:
\begin{equation}
     p_\mathcal{X}(x) = \frac{1}{N} \sum_{i=1}^N \mathsf{K} (\frac{\bb{x}-\bb{x}_i}{\sigma}),
\end{equation}
where $\sigma$ is the bandwidth parameter. We choose the Gaussian kernel $G_\sigma (\bb{x}, \bb{y})= \frac{1}{\sqrt{2 \pi} \sigma} \exp{(-\frac{\lVert \bb{x} - \bb{y}\rVert}{2 \sigma^2})}$ as the kernel function due to its nice properties: it is symmetric, positive definite, and its value approaches zero when the point $\bb{x}$ moves away from the center $\bb{y}$ while being controlled by a decay factor determined by $\sigma$.

Inspired by  \cite{tsin2004correlation,roy2007deformable,hasanbelliu2011robust,he2016self}, we adopt the  Cauchy-Schwarz (CS) divergence  \cite{jenssen2005optimizing}, denoted as $\mathcal{D}_{CS}(q,p)$, to measure the similarity between two density functions, \added[id=PAN, comment={}]{which is defined as
\begin{equation}
\begin{split}
    \mathcal{D}_{CS}(q,p)  & = - \log \Big( \frac{\int q(x) p(x) dx}{\sqrt{\int q(x)^2 dx \int p(x)^2 dx}} \Big)  \\
  & = - \log  \int  q(x) p(x) dx +  0.5 \log  \int q(x)^2 dx \\ &  \quad   +  0.5 \log \int p(x)^2 dx, 
\end{split}
\end{equation}}
which is symmetric for any two PDFs $q$ and $p$ such that $0 \le \mathcal{D}_{CS} < \infty$ where the minimum is obtained iff $q(x) = p(x)$. We substitute Gaussian kernel PDF estimators for $\mathcal{Y}$ and $\hat{\mathcal{Y}}$ into $q(x), p(x)$, and make straightforward manipulations based on the convolution theorem for Gaussian functions (see the detailed derivation in Appendix \ref{sec:cs}), which gives
\begin{equation}\label{eq:cs_divergence}
\begin{split}
    \mathcal{D}_{CS}  (p_\mathcal{\hat{Y}}, p_\mathcal{Y} )  &  = - \log \sum_{j=1}^N \sum_{i=1}^N  G_{\sqrt{2} \sigma} (\hat{\bb{y}}_j, \bb{y}_i) \\
    &  + 0.5 \log \sum_{j'=1}^N \sum_{j=1}^N G_{\sqrt{2} \sigma} (\hat{\bb{y}}_{j'}, \hat{\bb{y}}_j) \\ 
    & + 0.5 \log \sum_{i'=1}^N \sum_{i=1}^N G_{\sqrt{2} \sigma} (\bb{y}_{i'}, \bb{y}_i).
\end{split}
\end{equation}

Later, we will show that CS leads to better performance compared to CD and EMD objectives by handling outliers using the Gaussian kernels \cite{jenssen2006cauchy}. Specifically, the Gaussian kernel $G$ is able to mitigate the oversensitivity to outliers by suppressing large distances between reference and reconstructed shape points. In CD and EMD, these large distances due to outliers negatively impact model training, leading to degraded performance. It is worth mentioning that CS is closely related to graph cuts and Mercer kernel theory \cite{jenssen2006cauchy}.

\noindent \textbf{Implementation of the CS loss.} The CS divergence loss can be implemented in PyTorch with a few lines of code. To handle numerical issues, we leverage the Log-Sum-Exp trick as shown in Algorithm~\ref{alg:code}. 
\begin{algorithm}[hbt!]
\caption{The CS divergence implemented in PyTorch.}
\label{alg:code}
\definecolor{codeblue}{rgb}{0.25,0.5,0.5}
\lstset{
  backgroundcolor=\color{white},
  basicstyle=\fontsize{7.2pt}{7.2pt}\ttfamily\selectfont,
  columns=fullflexible,
  breaklines=true,
  captionpos=b,
  commentstyle=\fontsize{7.2pt}{7.2pt}\color{codeblue},
  keywordstyle=\fontsize{7.2pt}{7.2pt},
}
\begin{lstlisting}[language=python]
# reconstructed target and target point coordinates <-- B X N x 3 and B X N x 3
# bandwidth, the kernel bandwith <-- scalar
def gmm(rec_target, target, bandwidth):
    # B X N x N x 3  <-- B X N X 1 x C - B X 1 X N X C
    diff_ij = (rec_target.unsqueeze(2) - target.unsqueeze(1))
    # B X N x N
    factor = 2*bandwidth*bandwidth 
    # B X N x N
    diff_ij = (diff_ij**2).sum(-1).div(factor).mul(-0.5) - 0.5*math.log(2*math.pi) - math.log(math.sqrt(2)*bandwidth)
    
    dist = torch.logsumexp((diff_ij).reshape(diff_ij.shape[0], -1),dim=1).mean()
    
    return dist
    
def cs_divergence(rec_target, target, bandwidth):

    r_t_dist = -1 * gmm(rec_target, target, bandwidth)
    r_r_dist = 0.5 * gmm(rec_target, rec_target, bandwidth)
    t_t_dist = 0.5 * gmm(target, target, bandwidth)

    return r_t_dist + r_r_dist + t_t_dist
\end{lstlisting}
\end{algorithm}

\subsection{The Training Objective}

Similar to Equations~(\ref{eq:cross_reverse_weight}) and (\ref{eq:cross_reconstruction}) using $\boldsymbol{W}^{\mathcal{X}\mathcal{Y}}$ to reconstruct $\mathcal{F}^\mathcal{Y}$ and $\hat{\mathcal{Y}}$, 
we compute the reconstruction weights $\boldsymbol{W}^{\mathcal{Y}\mathcal{X}}$ to approximate the original input shape $\mathcal{X}$, which results in the reconstructed shape $\mathbb{R}^{N \times 3} \ni  \hat{\mathcal{X}} := [\hat{\bb{x}}_1, \dots, \hat{\bb{x}}_N]^T$. In addition, we approximate the original $\mathcal{Y}$ using $\boldsymbol{W}^{\mathcal{Y}\mathcal{Y}}$ and $\mathcal{Y}$, which is a self-construction process to obtain the approximate shape $\mathbb{R}^{N \times 3} \ni  \tilde{\mathcal{Y}} := [ \tilde{\bb{y}}_1, \dots, \tilde{\bb{y}}_N
]^T$. The approximation  of $\mathcal{X}$ is similarly expressed as $\mathbb{R}^{N \times 3} \ni  \tilde{\mathcal{X}} := [\tilde{\bb{x}}_1, \dots, \tilde{\bb{x}}_N]^T$. The final training objective is  defined as
\begin{equation}
\begin{split}
    \minimize_{\mathcal{F}}  & \quad \lambda_{\text{cross}} ( \mathcal{D}  (\mathcal{\hat{X}}, \mathcal{X} ) 
     +  \mathcal{D} (\mathcal{\hat{Y}} , \mathcal{Y} )) \\
    & +  \lambda_{\text{self}} ( \mathcal{D}  (\mathcal{\tilde{X}} ,  \mathcal{X} )  +   \mathcal{D} (\mathcal{\tilde{Y}} ,  \mathcal{Y}))  \\ 
    & + \lambda_{\text{reg}} (E_r(\mathcal{X}, \mathcal{\hat{Y}}) + E_r(\mathcal{Y}, \mathcal{\hat{X}}))
\end{split}
\end{equation}
where $\lambda_{\text{cross}}, \lambda_{\text{self}}$, and $\lambda_{\text{reg}}$ are the hyperparameters to balance different losses and $\mathcal{D} ( \cdot , \cdot)$ is the CS objective in Equation \ref{eq:cs_divergence}. $E_r( \cdot , \cdot)$ is the optional smoothness term defined as the mapping loss \cite{lang2021dpc}, which encourages points in $\hat{\mathcal{Y}}$ (or $\hat{\mathcal{X}}$) to remain close if their one-to-one corresponding points in $\mathcal{X}$ (or $\mathcal{Y}$) are close to each other. $E_r(\mathcal{X}, \mathcal{\hat{Y}})$ is defined as
\added[id=PAN, comment={}]{
\begin{equation}
\begin{split}
    E_r(\mathcal{X}, \mathcal{\hat{Y}})  = \frac{1}{N*K} \sum_{i=1}^N \sum_{l \in \mathcal{N}_{\mathcal{X}}( \bb{x}_i)} v_{i,l}^{\mathcal{X}\mathcal{\hat{Y}}} \Vert \hat{\bb{y}}_i - \hat{\bb{y}}_l \Vert_2^2
\end{split}
\end{equation}
where $\mathcal{N}_{\mathcal{X}}(\bb{x}_i)$ is the Euclidean neighborhood of $\bb{x}_i$ in $\mathcal{X}$ of size $K$, $v_{i,l}^{\mathcal{X}\mathcal{\hat{Y}}} = \exp{\frac{-\Vert \bb{x}_i - \bb{x}_l \Vert_2^2}{\alpha}}$ where $\alpha$ is a hyperparameter configured by following \cite{lang2021dpc}. $E_r(\mathcal{Y}, \mathcal{\hat{X}})$ is similarly defined.}

\subsection{Test Phase}

In the test phase, we obtain the correspondence for each source point $\bb{x}_i$ by selecting a point from the target shape whose embedding is the nearest neighbor of $\bb{x}_i$'s  embedding based on the cosine similarity. This gives
\begin{equation}\label{eq:nn}
    T_\mathcal{XY}(\bb{x}_i) = \bb{y}_{j^*}, \quad j^* = \argmax_{j \in \{1, \dots, N\}} \frac{(f^\mathcal{X}_i) \cdot (f^\mathcal{Y}_j)^T}{\Vert f^\mathcal{X}_i \Vert_2   \Vert f^\mathcal{Y}_j \Vert_2 }.
\end{equation}

{\noindent \textbf{Summary} We have presented the LTENet framework for unsupervised shape correspondence learning, which unifies nonlinear embeddings, LLE transformations in the embedding space, point cloud reconstruction, and implicit correspondence learning with the CS divergence. We consider the following analogy for LTENet: {\it CS divergences and top LLE transformations are to shape correspondence as Kullback–Leibler (KL) divergences and top linear classifiers are to classification. By doing so, we are able to learn universal feature embeddings where correspondences are directly obtained using nearest neighbors. This is also analogous to the open-set classification problem where we handle samples of unseen classes by comparing feature distances between these samples and nearest neighbor trained examples of seen classes.}

\section{Experiments}

In this section, we compare LTENet against recent state-of-the-art approaches on several well-established datasets for shape correspondence, and we conduct ablation studies.

\subsection{Experimental Setup}

\noindent \textbf{Datasets.} Following \cite{zeng2021corrnet3d,lang2021dpc}, we conduct experiments on standard datasets covering both human and nonhuman shapes. For human shapes, we use the large-scale SURREAL dataset \cite{groueix20183d} prepared by 3D-CODED \cite{groueix20183d}, which leverages  SMPL \cite{loper2015smpl} to generate a total of $230,000$ samples. We select arbitrary shapes as training pairs from SURREAL. We then evaluate on the challenging SHREC-19 \cite{melzi2019shrec} containing $430$ non-rigid shape pairs generated from $44$ real human scans. 
For non-human shapes, we adopt SMAL \cite{zuffi20173d} and TOSCA \cite{bronstein2008numerical} for training and evaluation, respectively. SMAL provides the 3D articulated parametric model for animals.
We create a training set of $10,000$ shapes by generating $2,000$ samples under each animal category. We pair arbitrary shapes of the same category. Similarly, we consider $41$ animal figures out of the total $80$ objects in TOSCA to match species in SMAL and generate $286$ test shape pairs from the SMAL dataset accordingly.

\begin{figure*}[t]
 \centering
 \begin{minipage}{0.33\linewidth}
 \centering
    \includegraphics[width=\textwidth]{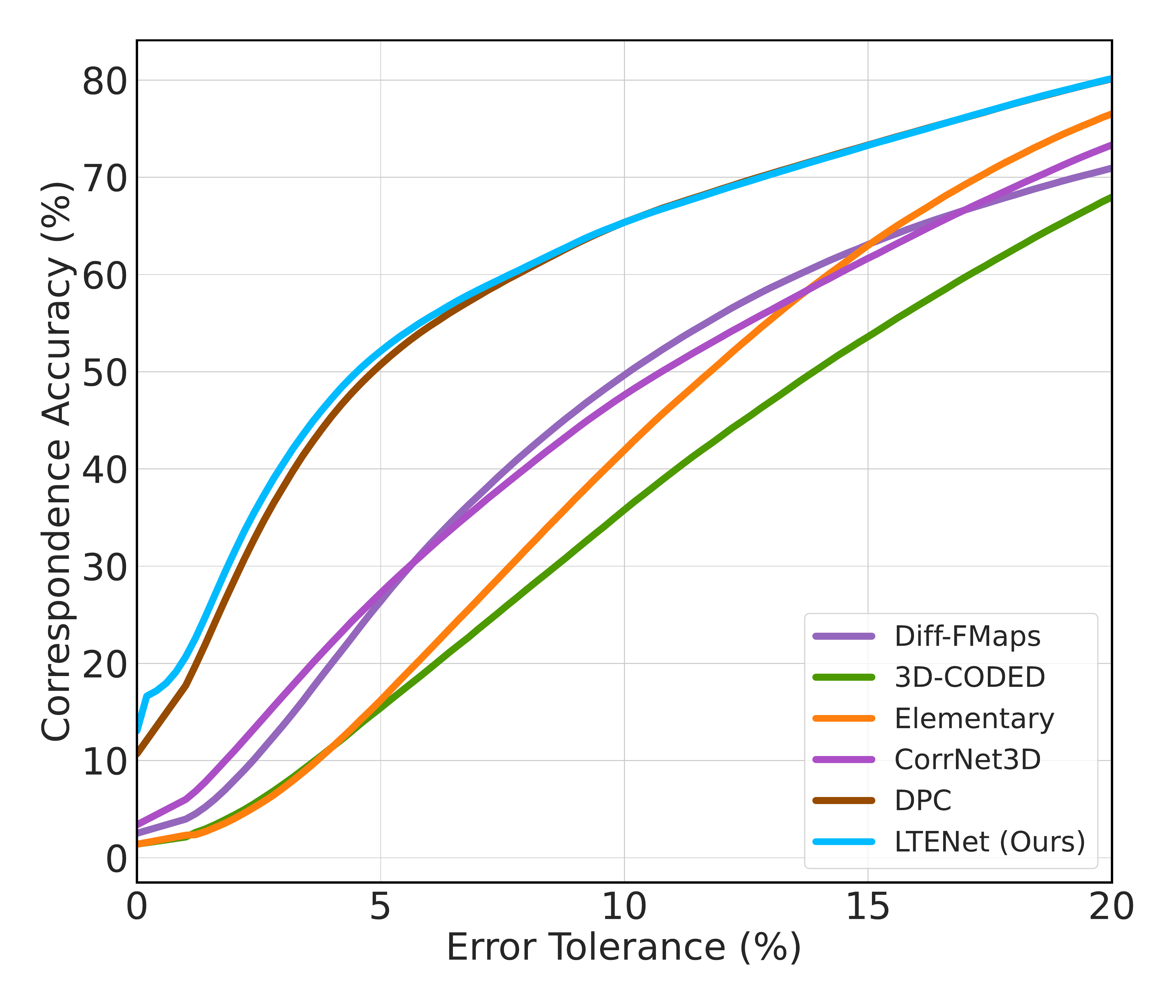}
  \end{minipage}
 \hfill
 \begin{minipage}{0.66\linewidth}
    \centering
    \scriptsize
    \begin{tabular}{cccc}
    Reference target & \whitetext{aaaaaaaaaaa}DPC \cite{lang2021dpc} & \whitetext{aaaaaaaaaaa} LTENet (ours) & \whitetext{a}Ground-truth \\
    \multicolumn{4}{c}{\includegraphics[width=\textwidth]{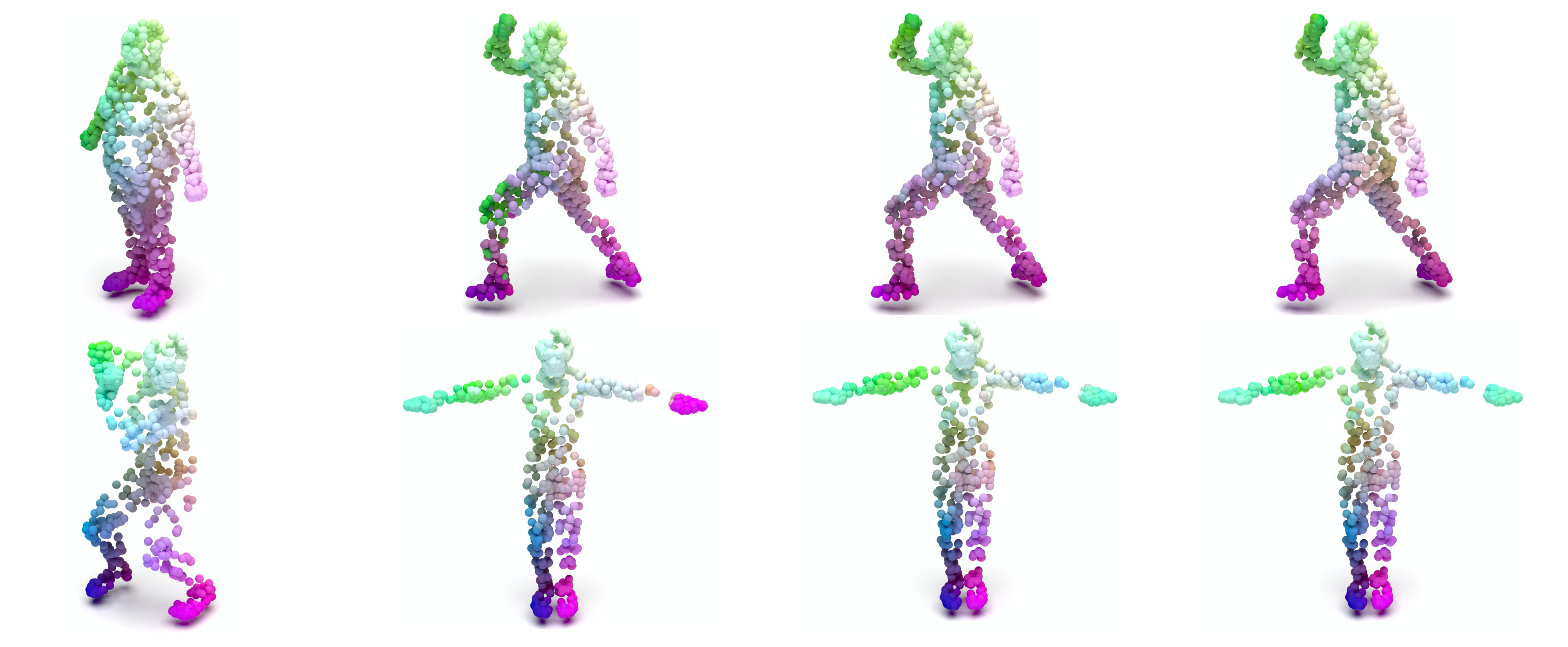}} 
    \end{tabular} 
  \end{minipage}
  \caption{(Left) \textbf{The correspondence accuracies under different error tolerance values in the SURREAL/SHREC setting}. Our method achieves better performance compared to the state-of-the-art DPC. (Right) \textbf{Visual examples of SHREC test pairs}. DPC contains outlier matches, e.g., wrongly matching hands to thighs or feet to hands.  LTENet generates more accurate and smoother predictions.}\label{fig:train_surreal_test_shrec}
\end{figure*}

\noindent \textbf{Evaluation metrics.} 
A common evaluation metric is the geodesic distance error assuming a known point adjacency matrix, which is unavailable in point clouds. Instead, we follow
\cite{lang2021dpc} to calculate the  correspondence error as
\begin{equation}
    err = \frac{1}{N} \sum_{i=1}^N \lVert T_\mathcal{XY}(\bb{x}_i) - T_\mathcal{XY}^{gt}(\bb{x}_i) \rVert_2.
\end{equation}
where $T_\mathcal{XY}(\bb{x}_i), T_\mathcal{XY}^{gt}(\bb{x}_i)$ denote the predicted and ground truth correspondence of point $\bb{x}_i$ w.r.t. $\mathcal{Y}$ and $\lVert \cdot \rVert_2$ is the $\ell_2$ norm of a vector. Additionally, we use the error tolerance $\epsilon = r/dist_{max}$ coupled with a tolerant radius $r$, where $dist_{max} = \max \{ \lVert \bb{y}_i - \bb{y}_j \rVert_2, \forall i, \forall j\}$ denotes the maximal distance of all pairwise point distances in $\mathcal{Y}$. The correspondence accuracy under $\epsilon$ is defined as 
\begin{equation}
    acc(\epsilon) = \frac{1}{N} \sum_{i=1}^N \boldsymbol{\mathbbm{1}} (\lVert T_\mathcal{XY}(\bb{x}_i) - T_\mathcal{XY}^{gt}(\bb{x}_i) \rVert_2 < \epsilon * dist_{max}),
\end{equation}
where $\boldsymbol{\mathbbm{1}}$ is the indicator function. We set different $\epsilon$ values between $0\%$ to $20\%$.

\noindent \textbf{Implementation details.} The proposed LTENet is not limited to a specific model architecture for the embedding network $\mathcal{F}$. We followed \cite{zeng2021corrnet3d,lang2021dpc} to adopt a variant of DGCNN \cite{wang2019dynamic} as $\mathcal{F}$, where its core component is the popular EdgeConv operator that builds a dynamic graph over points for learning the feature embeddings.
We refer the reader to \cite{wang2019dynamic} for more details. Our models were implemented in Pytorch \cite{paszke2019pytorch}. We used the AdamW optimizer \cite{loshchilov2017decoupled} with an initial learning rate of $0.0003$,  momentum $0.9$, and weight decay of $0.0005$. We used a cosine decay learning rate scheduler for 300 epochs and 10 epochs of linear warm-up. We trained models with a batch size of $8$ on a server equipped with AMD EPYC ROME microprocessors and NVIDIA A100 GPUs.

\noindent \textbf{Baseline methods.} We consider recent state-of-the-art unsupervised shape correspondence learning  approaches (DPC \cite{lang2021dpc} and CorrNet3D \cite{zeng2021corrnet3d}) as competitive baselines. We compare against supervised approaches, including Diff-FMaps \cite{marin2020correspondence}, 3D-CODED \cite{groueix20183d}, and Elementary Structures \cite{deprelle2019learning}, and mesh-based approaches, including  
the unsupervised SURFMNet \cite{roufosse2019unsupervised} and the supervised GeoFMNet \cite{donati2020deep}.

\subsection{Results on Human Datasets} 

We explored two training and evaluation settings. For a fair comparison, we followed DPC \cite{lang2021dpc} to train our models by selecting the first 2000 shapes out of the total 230,000 samples in SURREAL and evaluated  on the official $430$ SHREC pairs (SURREAL/SHREC). We also trained models on random pairs generated from SHREC and evaluated on the same test pairs (SHREC/SHREC).

\begin{figure*}[t]
  \begin{minipage}{0.33\linewidth} 
  \centering
    \includegraphics[width=\textwidth]{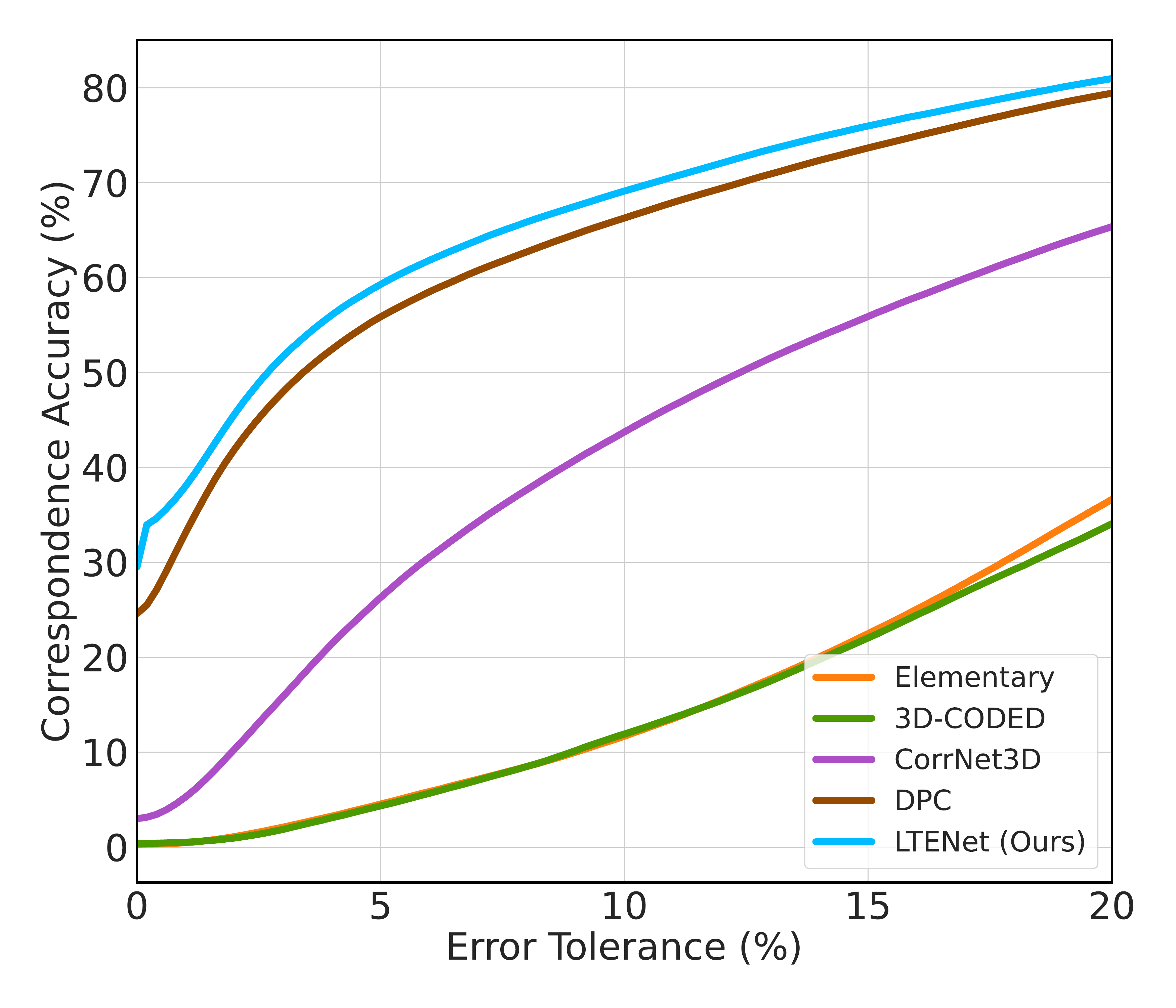}
  \end{minipage}
 \hfill
  \begin{minipage}{0.66\linewidth}
    \centering
    \scriptsize
    \begin{tabular}{cccc}
    Reference target & \whitetext{aaaaaaaaaaaa}DPC \cite{lang2021dpc} & \whitetext{aaaaaaaaaa} LTENet (ours) & \whitetext{a}  Ground-truth \\
    \multicolumn{4}{c}{\includegraphics[width=\textwidth]{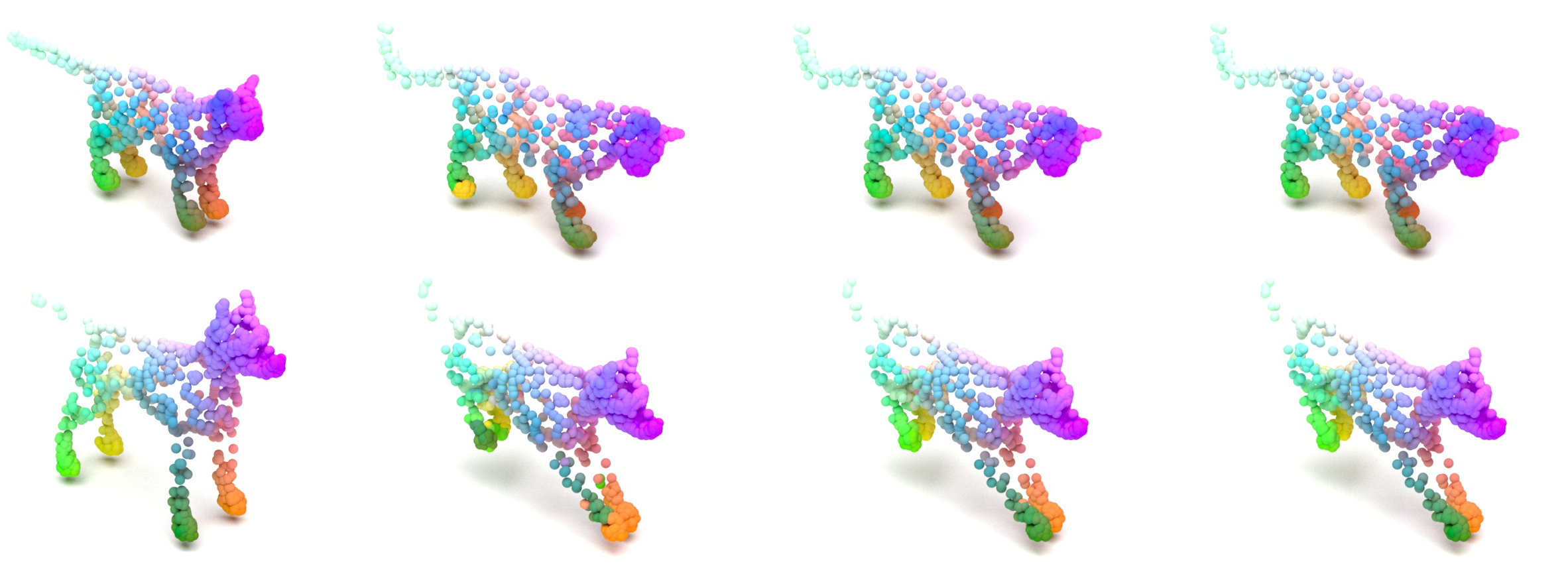}}
    \end{tabular} 
  \end{minipage}
  \caption{(Left) \textbf{The correspondence accuracies under different error tolerance values in the SMAL/TOSCA setting}. Our method substantially improves the correspondence accuracies under all tolerance values. (Right) \textbf{Visual examples of TOSCA test pairs}. DPC suffers from prediction errors caused by the difficulty of distinguishing between left and right or rear and front legs. Our method generates more accurate correspondence predictions closer to the ground truth correspondence maps.}\label{fig:train_smal_test_tosca}
\end{figure*}

\begin{table}[t]
\caption{\textbf{Accuracy and error.} The proposed LTENet achieves state-of-the-art shape correspondence performance, indicated by the correspondence accuracy at 1\% tolerance (\textit{acc}, in percentage) and the average correspondence error (\textit{err}, in centimeters).}
\centering
\label{tab:main}
\resizebox{.48\textwidth}{!}{
\begin{tabular}{lcccccc}
\toprule
       & \multicolumn{2}{c}{\textbf{SURREAL/}} & \multicolumn{2}{c}{\textbf{SHREC/}}  & \multicolumn{2}{c}{\textbf{SMAL/}} \\
      & \multicolumn{2}{c}{\textbf{SHREC}}    & \multicolumn{2}{c}{\textbf{SHREC/}}  & \multicolumn{2}{c}{\textbf{TOSCA}}   \\
       \cmidrule(lr){2-3} \cmidrule(lr){4-5} \cmidrule(lr){6-7} 
Method & \textit{acc} $\uparrow$ & \textit{err} $\downarrow$ & \textit{acc} $\uparrow$ & \textit{err} $\downarrow \hspace{0.5em}$  & \textit{acc} $\uparrow$ & \textit{err} $\downarrow \hspace{0.5em}$ \\

\midrule

SURFMNet~\cite{roufosse2019unsupervised}  & 4.3\% & 0.3 & 5.9\% & 0.2 & * & *  \\
GeoFMNet~\cite{donati2020deep}  & 8.2\% & 0.2 & * & * & * & *   \\

\midrule
Diff-FMaps~\cite{marin2020correspondence}  & 4.0\% & 7.1 & * & *   & * & *     \\
3D-CODED~\cite{groueix20183d}     & 2.1\% & 8.1 & * & *       & 0.5\% & 19.2  \\
Elementary~\cite{deprelle2019learning}  & 2.3\% & 7.6 & * & *   & 0.5\% & 13.7   \\
\midrule
CorrNet3D~\cite{zeng2021corrnet3d}   & 6.0\% & 6.9 & 0.4\% & 33.8 & 5.3\% & 9.8  \\
DPC \cite{lang2021dpc}   & 17.7\% & 6.1 & 
\textbf{15.3\%} & 5.6 & 33.2\% & 5.8   \\
Released DPC \cite{lang2021dpc}   & 17.5\% & 6.3 & 14.5\% & \textbf{5.3} & 33.5\% & 5.8  \\
\midrule
LTENet (Ours)  & \textbf{20.7\%} & \textbf{5.8} & 14.3\% & 6.0 & \textbf{38.1\%}  & \textbf{5.7} \\

\bottomrule
\end{tabular}
}
\end{table}

\noindent \textbf{Quantitative evaluation.}   Table \ref{tab:main} summarizes the $\textit{acc}$ at $1\%$ error tolerance, indicating a near-perfect correspondence matching and average correspondence error $\textit{err}$. On SHREC/SHREC, we achieved a competitive performance against DPC. On SURREAL/SHREC, we outperformed all baseline models.  Specifically, SURFMNet~\cite{roufosse2019unsupervised} and GeoFMNet~\cite{donati2020deep} achieved impressive performance. However, they require the expensive computation of the LBO basis and complex test-time post-processing \cite{Simone2019zoomout,besl1992method}. Our method achieved approximately $5 \times$ and $2.5 \times$ accuracies compared to SURFMNet and GeoFMNet, respectively, while showing a comparable run-time
inference speed to DPC \cite{lang2021dpc}, which is about $100 \times$ faster against SURFMNet and GeoFMNet. Diff-FMaps \cite{marin2020correspondence} suffers from over-fitting on training samples without exploiting shape priors, e.g., local smoothness. CorrNet3D \cite{zeng2021corrnet3d} shows improvements over 3D-CODED \cite{groueix20183d} and Elementary Structures \cite{deprelle2019learning} but requires nontrivial optimization in the Sinkhorn-inspired DeSmooth module and the decoder, which limits its generalization performance. DPC \cite{lang2021dpc} is the current state-of-the-art method using learning
the latent affinity via a simplified point reconstruction. Our LTENet achieved the best $\textit{acc}$ of $20.7\%$ and the lowest $\textit{err}$ of $5.8$, which significantly exceeds the accuracy of DPC by $17.0\%$ and reduces the error by $4.9\%$. The accuracies in Figure \ref{fig:train_surreal_test_shrec} (left) indicate that we achieved a clear improvement, especially for almost-perfect matching with $\epsilon < 5\%$.

\noindent \textbf{Qualitative evaluation.} We provide visual examples in Figure \ref{fig:train_surreal_test_shrec} (right), showing the clear improvement made by LTENet (see more results in the appendix).

\subsection{Results on the Nonhuman Datasets}

We trained models on the SMAL dataset and evaluated on the unseen TOSCA dataset that contains animal objects with diverse poses (SMAL/TOSCA).
\begin{figure*}
  \begin{minipage}{0.33\linewidth} 
  \centering
    \includegraphics[width=\textwidth]{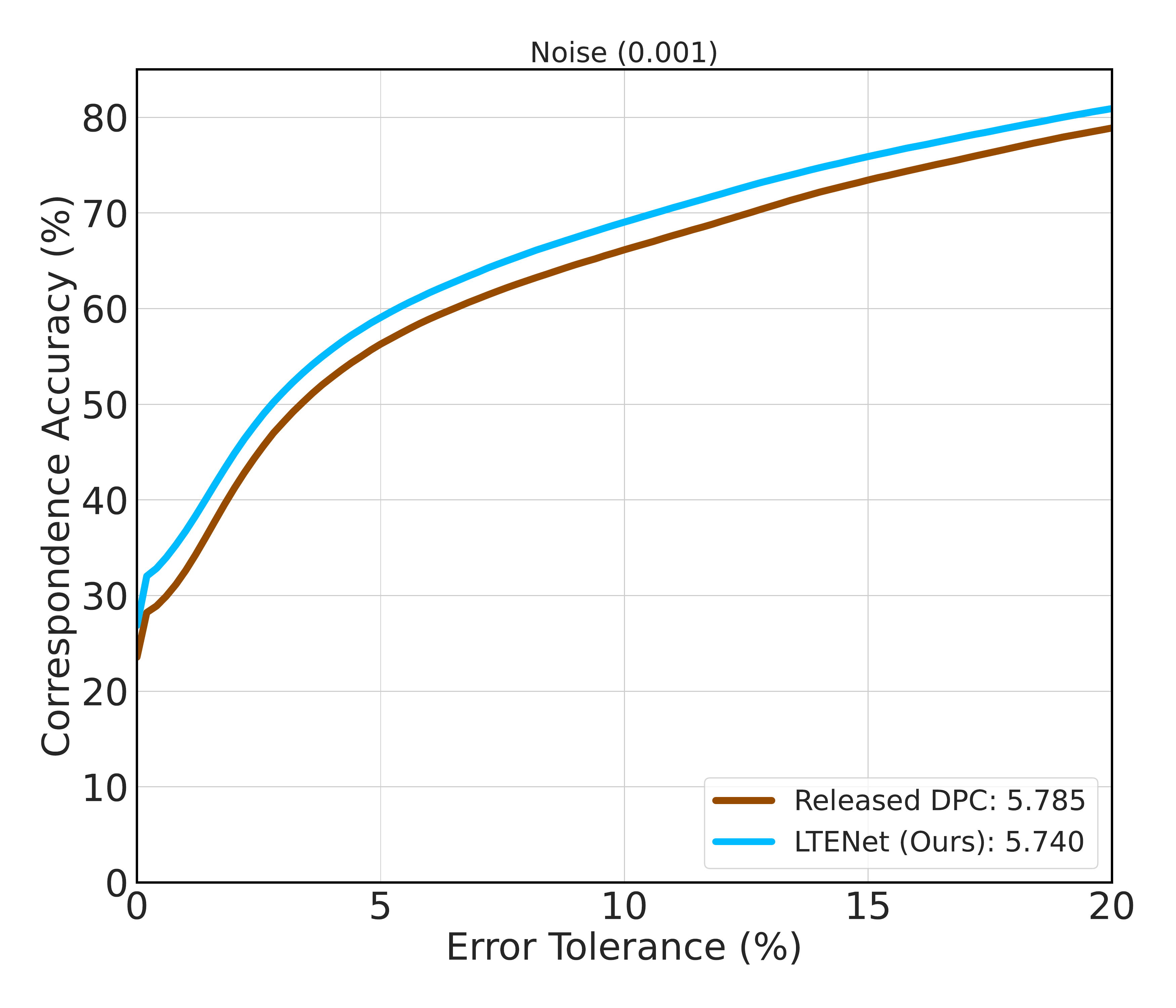}
    (a)
  \end{minipage}
 \hfill
  \begin{minipage}{0.33\linewidth}
  \centering
    \includegraphics[width=\textwidth]{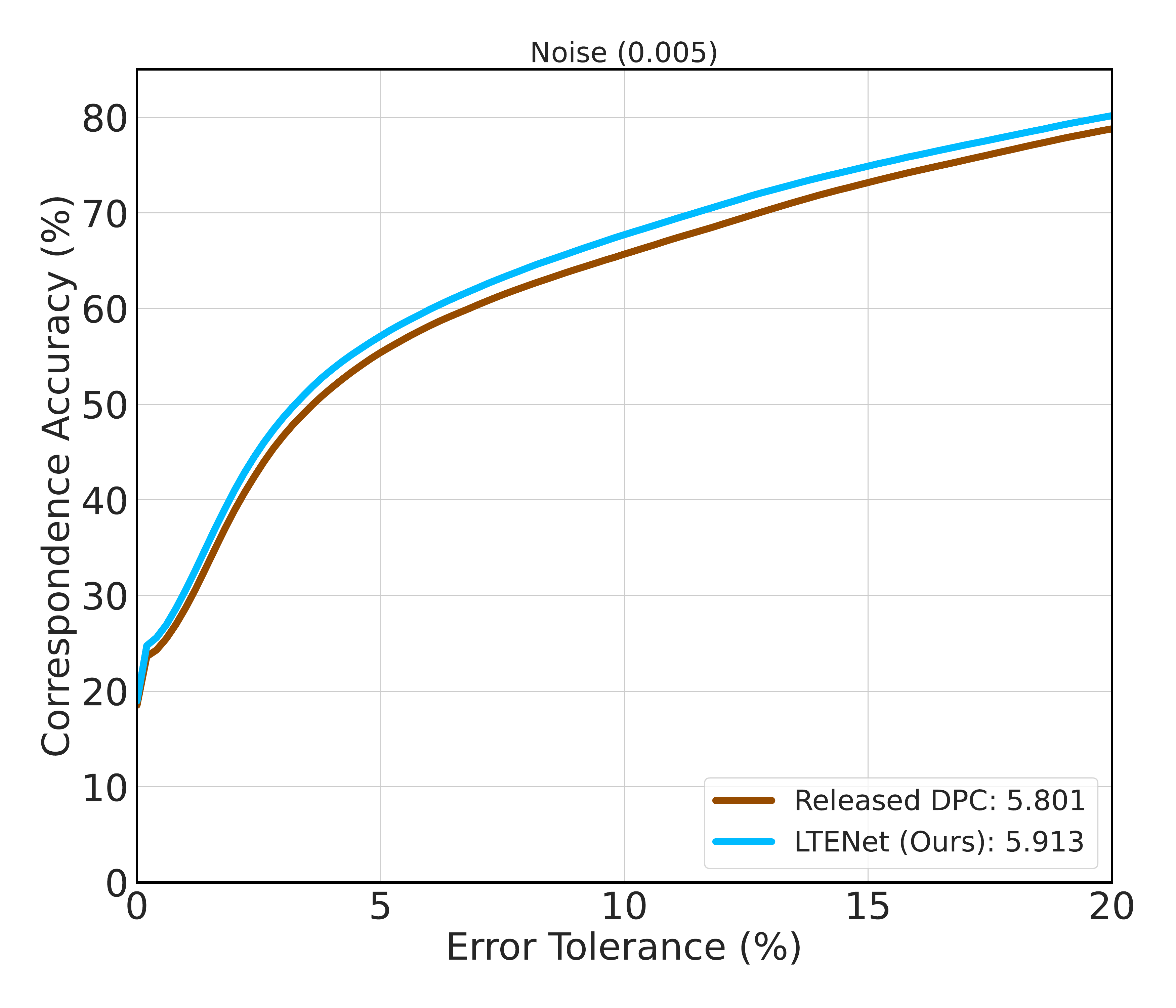}
    (b)
  \end{minipage}
  \hfill
  \begin{minipage}{0.33\linewidth}
  \centering
    \includegraphics[width=\textwidth]{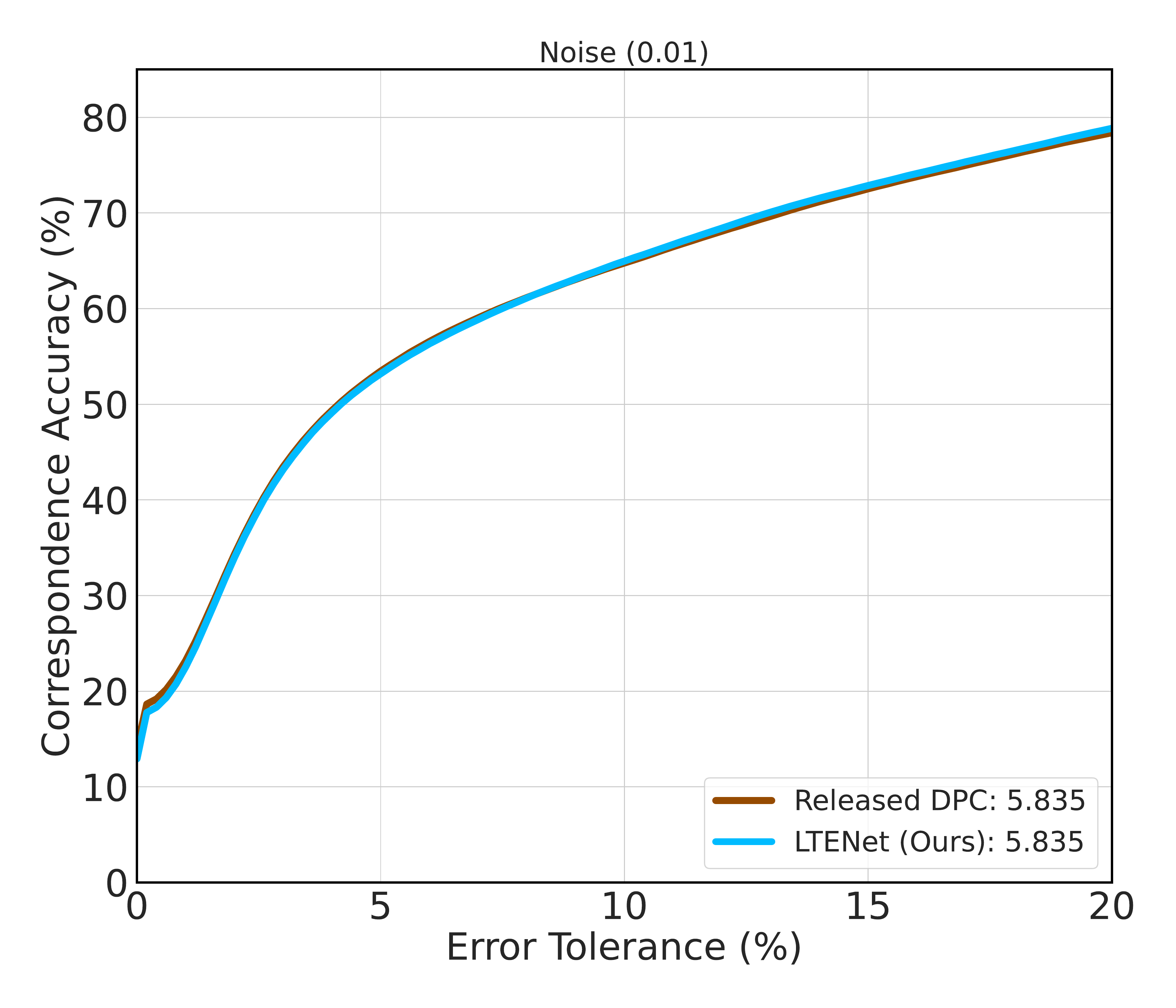}
    (c)
  \end{minipage}
  \caption{The evaluation of correspondence prediction of TOSCA test point clouds in the SMAL/TOSCA setting with additional noise. From (a) to (c), we gradually add stronger Gaussian noise with zero means and larger standard deviations, i.e., $0.001, 0.005, 0.01$, to source shapes. Our method
demonstrates its resilience against noise.}\label{fig:noise_tosca}
\end{figure*}

\noindent \textbf{Quantitative evaluation.} As shown in Table \ref{tab:main}, LTENet achieved the best performance on SMAL/TOSCA in terms of the $\textit{acc}$ at $1\%$ and $\textit{err}$. The significant pose and shape differences between SMAL and TOSCA impact 3D-CODED and Elementary Structures relying on a single standard template, e.g., a standing cat. They struggle to handle shapes in different categories and various poses in the TOSCA test pairs. The proposed LTENet achieves an $\textit{acc}$ of $38.1\%$ at $1\%$ error, obtaining an increase of $4.3\%$ in absolute accuracy compared to DPC's best $\textit{acc}$ of $33.8\%$. The detailed correspondence accuracy under different error tolerance values can be found in Figure \ref{fig:train_smal_test_tosca} (left), which shows that our method obtains a substantial improvement over other methods.

\noindent \textbf{Qualitative evaluation.}  Figure \ref{fig:train_smal_test_tosca} (right) provides some visual results on the TOSCA test pairs, which verify that our method could generate more accurate correspondence predictions. More results can be found in the appendix.

\subsection{Model Robustness under Presence of Noise}

\added[id=PAN,comment={}]{We investigate the robustness of the learned embeddings by perturbing the test dataset with Gaussian noise in the setting of SMAL/TOSCA, which is particularly challenging due to the presence of noise that ruins the underlying shape structure. Specifically, we select DPC as the competitive baseline. For the test samples from the TOSCA dataset, we add Gaussian noise with zero means and different standard deviations, i.e., $0.001, 0.005, 0.01$ to source shapes.}

\added[id=PAN,comment={}]{As can be seen in Figure \ref{fig:noise_tosca}, Our approach outperforms the state-of-the-art DPC  in terms of correspondence accuracy and shows comparable performance in correspondence errors. Our method demonstrates moderate resilience against noise.  Added noise heavily drops those correspondence accuracies under small error tolerance, e.g., less than $5 \%$. In a similar manner, the additional experiment conducted in the setting of SURREAL/SHREC can be found in the appendix.}

\begin{table}
\centering
\caption{\textbf{The benefit of pursuing local linearity for embedding learning}. We conduct the ablation
study under the SURREAL/SHREC setting. Our LTENet framework implicitly imposes a smooth constraint on embeddings such that it performs competitively without a regularization, while DPC \cite{lang2021dpc} suffers a significant performance drop. }\label{tab:ablation}
\begin{tabular}{lll}
\toprule
Method & \textit{acc} $\uparrow$ & \textit{err} $\downarrow \hspace{0.5em}$ \\
\midrule
DPC (without $E_r$)  & 11.4\% & 6.7 \\
DPC  & 17.7\%   (+6.3\%) & 6.1  (-0.6) \\
LTENet (without $E_r$)  &  20.6\% (+9.2\%)  & 5.9 (-0.8) \\
LTENet  & 20.7\%  (+9.3\%) & 5.8  (-0.9) \\
\bottomrule
\end{tabular}
\end{table}

\begin{figure*}[t]
  \begin{minipage}{0.33\linewidth} 
  \centering
    \includegraphics[width=\textwidth]{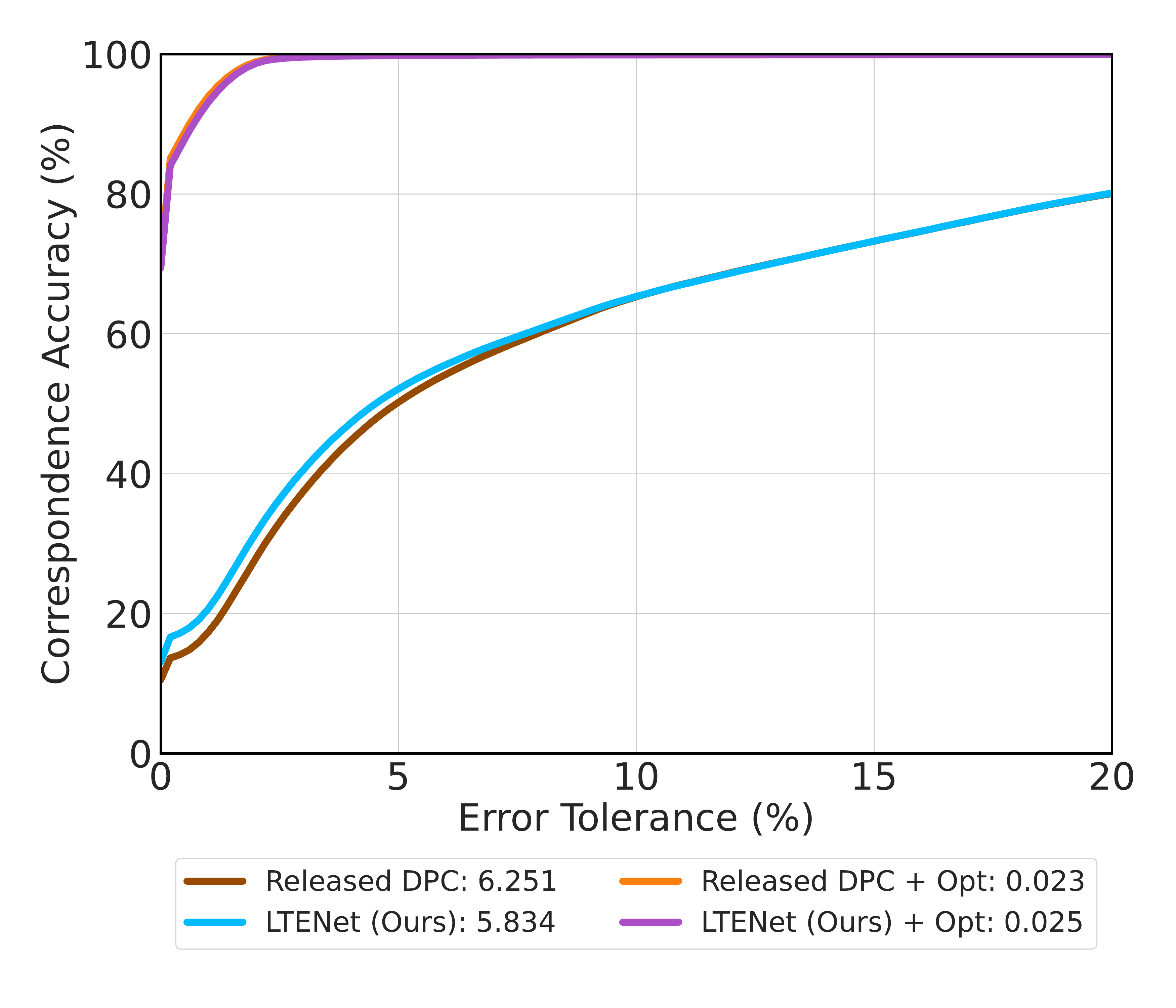}
    (a)
  \end{minipage}
  \hfill
  \begin{minipage}{0.33\linewidth}
  \centering
    \includegraphics[width=\textwidth]{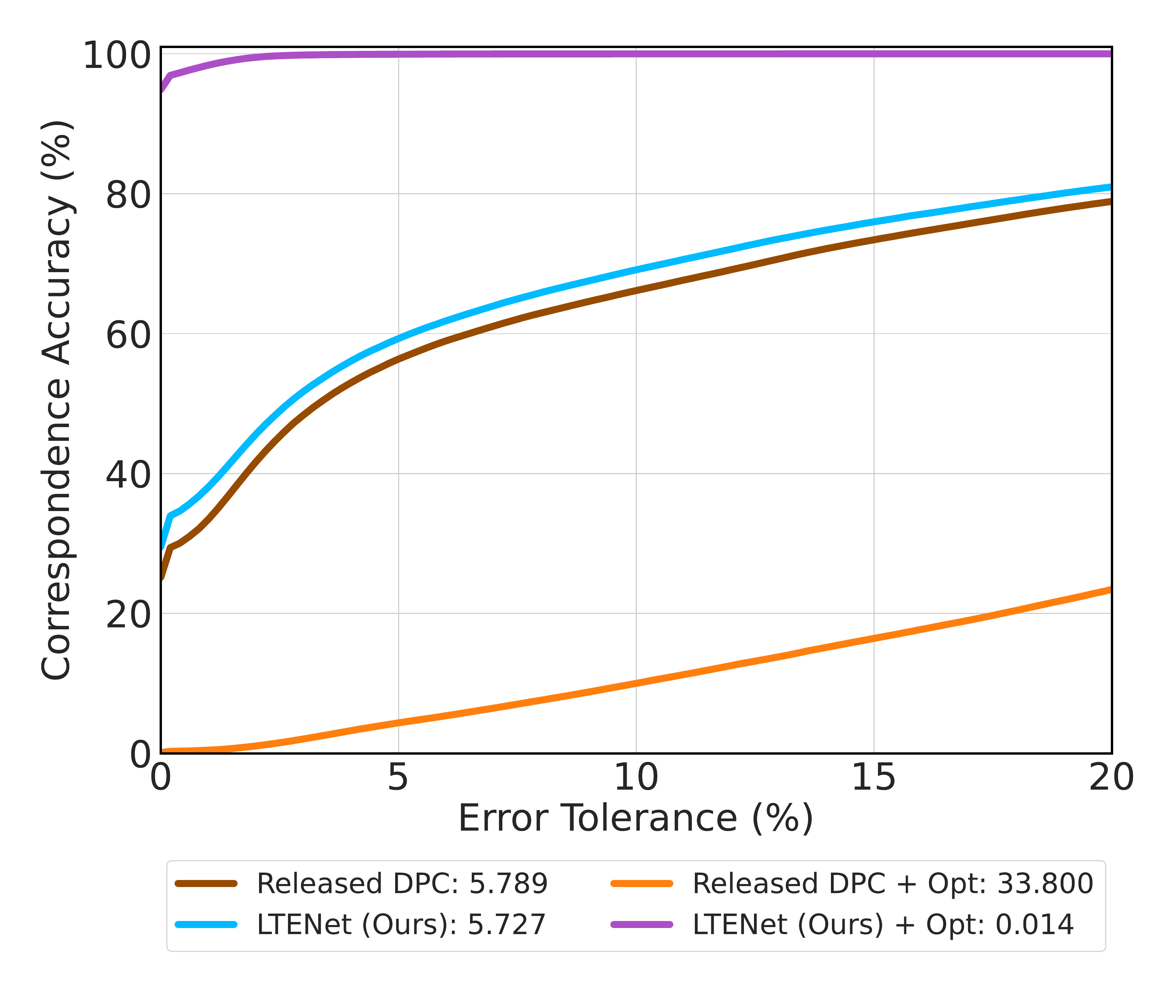}
    (b)
  \end{minipage}
  \hfill
  \begin{minipage}{0.33\linewidth}
  \centering
    \includegraphics[width=\textwidth]{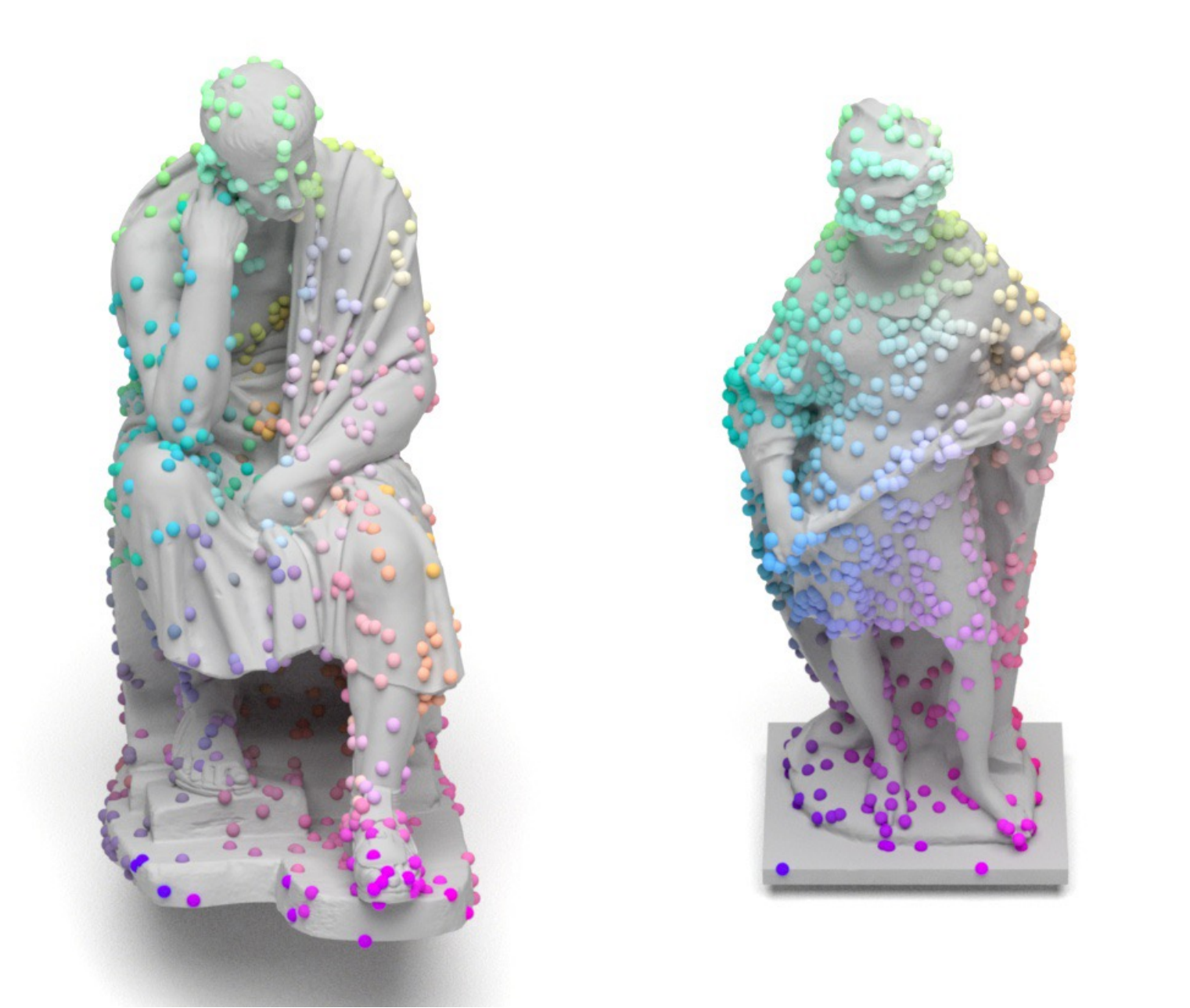}
    (c)
  \end{minipage}
  \caption{(a) The performance gap between the learned embeddings from our approach and
the transformed embeddings using optimal linear transformations in the SURREAL/SHREC setting. Adding the optimal linear transformation to learned embeddings significantly boosts the performance of DPC and our approach and leads to nearly perfect correspondence matching. (b) The performance gap evaluated similarly in the SMAL/TOSCA setting. Adding the optimal linear transformation is beneficial to our approach while being harmful to DPC, which suggests that a potential mismatch problem might have happened --- the embedding obtained from DPC is suitable for shape correspondence, but it is not necessarily suitable for use as the basis in a functional map framework where shape embeddings are related by a linear transformation.
 (c) The qualitative example of correspondence predictions between a pair of real-world scans. Despite the significant differences in pose and the presence of challenging non-isometry, our approach shows its resilience by producing reliable correspondence results.}\label{fig:global_transformation}
 \vspace{2mm}
\end{figure*}
\subsection{Comparisons between DPC and LTENet}

\added[id=PAN,comment={}]{Our LTENet highlights \textit{a novel approach to learning \textbf{locally linear} shape embeddings capable of capturing the underlying
structure of the shape manifold}, which we achieve by marrying LLE with the construction of high-dimensional neighborhood-preserving shape embeddings. We built our architecture following the self- and cross-reconstruction framework in DPC \cite{lang2021dpc}. Though both seek to learn good embeddings, our LTENet encourages the best \textbf{locally linear} alignment between shape embeddings without ambiguity via the closed-form expression of reconstruction weights.}

\added[id=PAN,comment={}]{Table~\ref{tab:ablation} demonstrates its benefit by summarizing our key results and additional results from the appendix of DPC. We clarify that $E_r$ (the smoothness term) is the same mapping loss as in DPC. Due to DPC's lack of a mechanism to enforce local linearity of embeddings, $E_r$ is required in DPC for better performance. Without this term, DPC suffers a significant drop from an \textit{acc} of $17.7\%$ to $11.4\%$. Our LTENet significantly outperforms DPC by enforcing a suitable manifold learning on shape correspondence under the same setting.
}

\subsection{Combining Locally Linear Embeddings with Functional Map-inspired Globally Linear Transformations}

\added[id=PAN,comment={}]{An interesting exploration is to understand the performance gap between the learned embeddings from our LTENet and their optimal embeddings transformed from globally linear transformations using the ground-truth correspondence. Inspired by the functional map paradigm \cite{ovsjanikov2012functional,marin2020correspondence} where shape
embeddings are related by a linear transformation, given the fixed point-to-point correspondence matrix $\Pi_\mathcal{XY}$ and a pair of shapes $\mathcal{X}$ and $\mathcal{Y}$, we treat our learned embedding $\mathcal{F}^\mathcal{X}, \mathcal{F}^\mathcal{Y}$ as the fixed bases and retrieve the optimal linear transformation $\mathcal{A}_{\mathcal{XY}}  = ((\mathcal{F}^\mathcal{X})^\dagger \Pi_\mathcal{XY} \mathcal{F}^\mathcal{Y})^T$ (see the detailed derivation in Appendix \ref{sec:optimal}). Under both settings of SURREAL/SHREC and SMAL/TOSCA, we evaluate (1) the matching using the learned embeddings $\mathcal{F}^\mathcal{X}$ and $\mathcal{F}^\mathcal{Y}$, (2) the matching estimated by finding  nearest neighbors between $\mathcal{F}^\mathcal{X} \mathcal{A}_{\mathcal{XY}}^T$ and $\mathcal{F}^\mathcal{Y}$, and (3) the matching by replacing our embeddings with DPC's embeddings and following the evaluation in (1) and (2).}

\begin{figure*}[t]
  \begin{minipage}[t]{0.49\linewidth}
    \includegraphics[width=\textwidth]{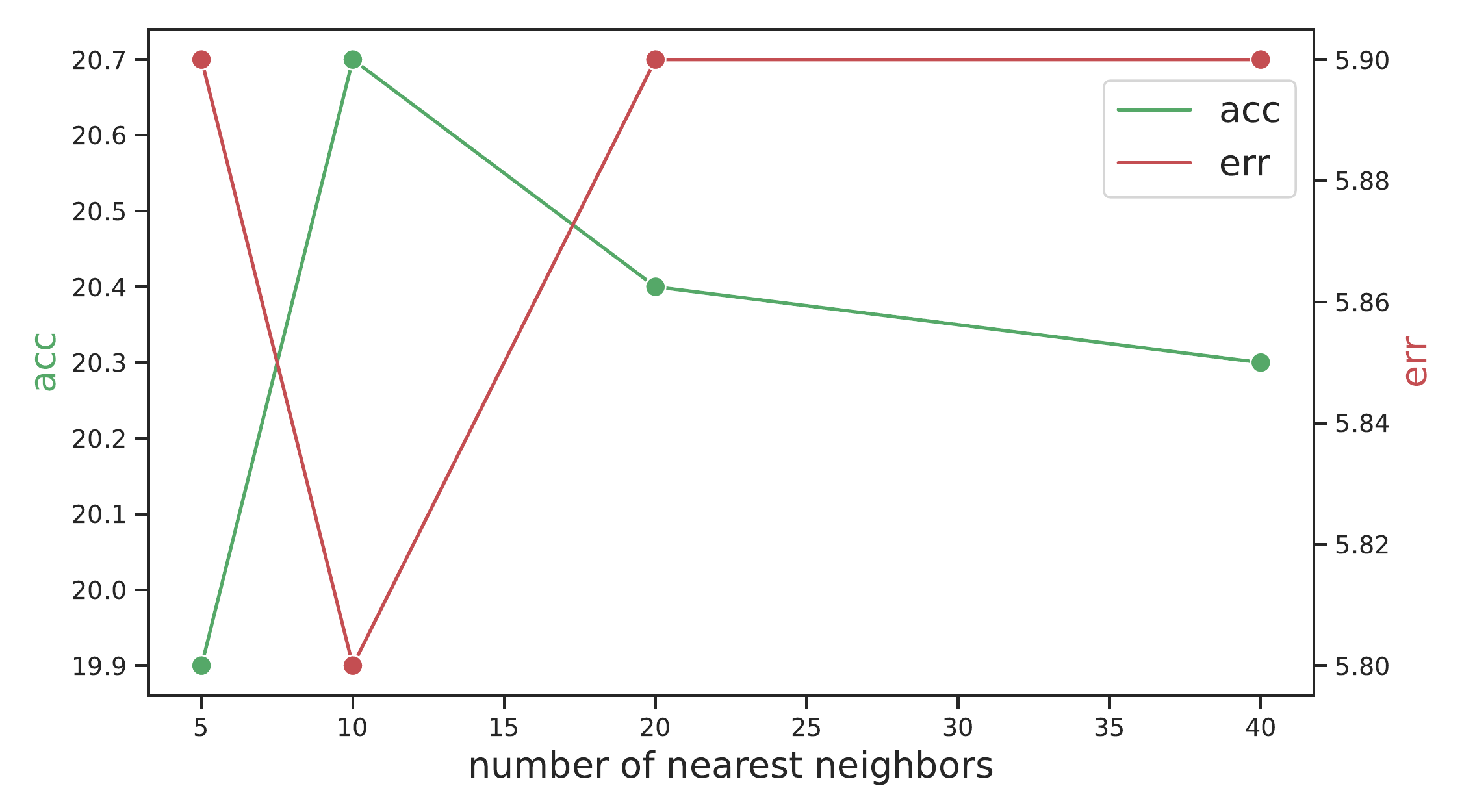}
  \end{minipage}
 \hfill
  \begin{minipage}[t]{0.49\linewidth}
    \includegraphics[width=\textwidth]{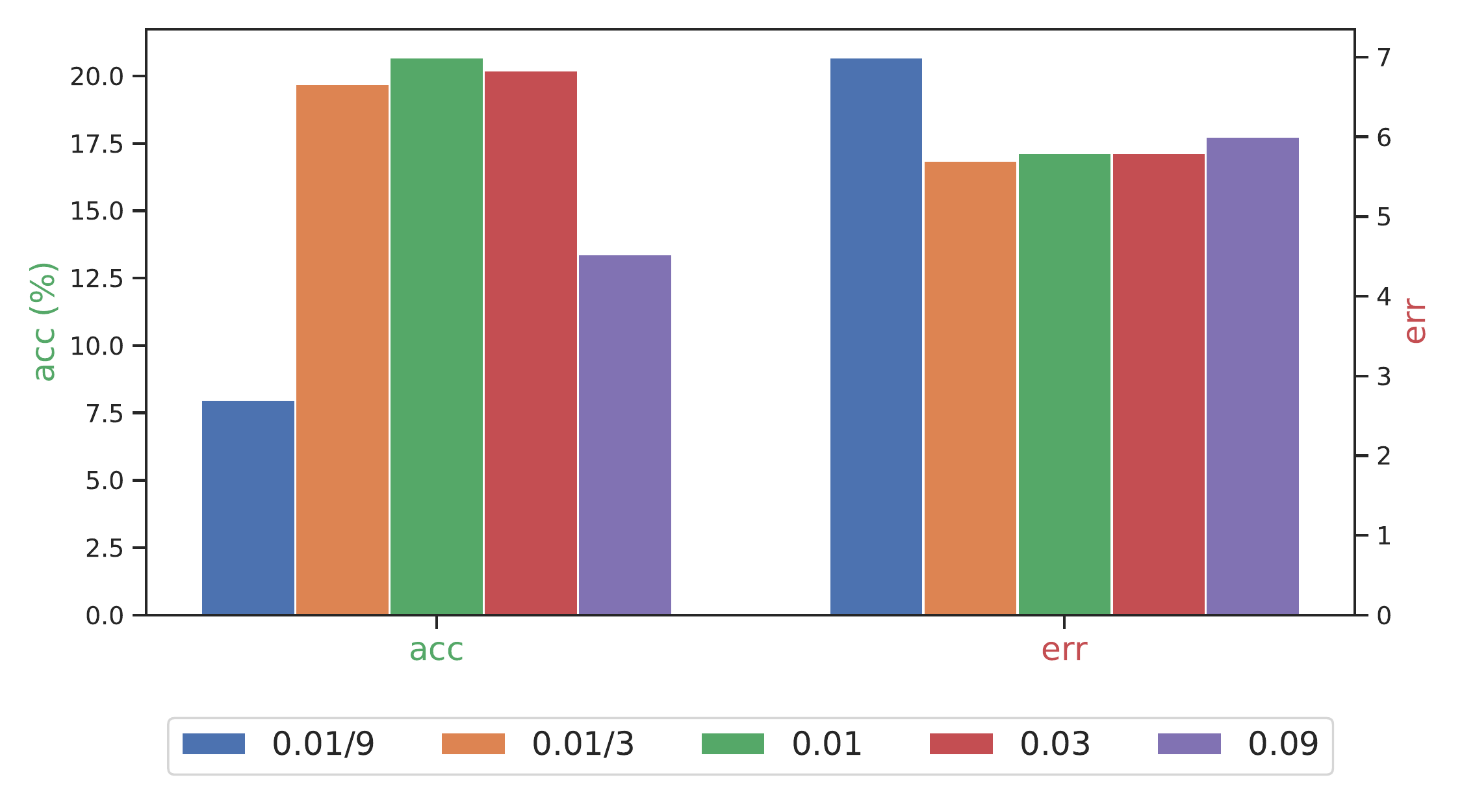}
  \end{minipage}
  \caption{(left) \textbf{Model performance with different numbers of nearest neighbors;} (right) \textbf{Model performance with different kernel bandwidths.} Choosing a suitable bandwidth or number of nearest neighbors leads to better performance.}    \label{fig:knn_bandwidth}
\end{figure*}
\added[id=PAN,comment={}]{The results are summarized in Figure \ref{fig:global_transformation}. Interestingly, we observe consistent improvements after applying the additional linear transformation (Opt) to LTENet in both SURREAL/SHREC and SMAL/TOSCA.  DPC achieves better performance in SURREAL/SHREC with optimal linear transformation. However, it suffers a significant performance drop in SMAL/TOSCA. This suggests that a potential mismatch problem might have happened --- the embedding obtained from DPC is suitable for shape correspondence, but it is not necessarily suitable to be used as the basis in a functional map framework. Accordingly, it implies that shape embeddings of DPC are not associated with the linear transformation in its  higher-dimensional embedding space. Though our approach does not suffer from such a mismatch problem, we admit that only a limited study was conducted on standard datasets covering human and nonhuman shapes. Future work is suggested to further investigate the mismatch problem by harmonizing  universal embeddings and linear transformations, ideally in an unsupervised learning setting, which is promising given the great progress made in unsupervised functional map learning \cite{halimi2019unsupervised,roufosse2019unsupervised,ginzburg2020cyclic}.}

\begin{table}
\centering
\caption{\textbf{Ablation study on model choices}. We conduct the ablation study under the SURREAL/SHREC setting.}
\label{tab:ablation_study}
\large
\resizebox{0.49\textwidth}{!}{%
\begin{tabular}{lccccc}
\toprule
Method & $\lambda_{\text{self}}=1$ &  $\lambda_{\text{cross}}=1$ & $\lambda_{\text{reg}}=10$ & \textit{acc} $\uparrow$ & \textit{err} $\downarrow \hspace{0.5em}$ \\
\midrule
Ours (Self) & \cmark &  &   & 2.3\% & 8.9  \\
Ours (Cross) &  & \cmark &  & 20.6\% & 6.5  \\
Ours (Self \& Cross) & \cmark & \cmark &  & 20.6\%  & 5.9 \\

\midrule
Ours (CD) & \cmark & \cmark & \cmark & 13.9\% & 6.3  \\
Ours (EMD) & \cmark & \cmark & \cmark & 19.6\% & 6.0  \\
\midrule
Ours &  \cmark & \cmark & \cmark  & 20.7\% & 5.8 \\ 

\bottomrule
\end{tabular}
}
\end{table}

\subsection{Evaluation on Real-world Data}

\added[id=PAN,comment={}]{
Inspired by \cite{zeng2021corrnet3d,lang2021dpc}, we examine our approach on model generalization and robustness by visualizing correspondence predictions between a pair of real-scanned shapes from the \textit{Scan the World} project
collection \cite{scan_the_world} in Figure \ref{fig:global_transformation} (c). We create the test point clouds by randomly sampling $1024$ points from each original 3D shape model. Despite the fact that these two shapes have significant differences in pose and present challenging non-isometry, our approach shows its resilience by producing reliable correspondence results. 
}

\subsection{Ablation Study}

We conduct the ablation study to evaluate the comparative effectiveness of the different components in LTENet under controlled experiments. All ablated versions were trained and evaluated following the SURREAL/SHREC setting.

\noindent \textbf{Choices of the model designs.} We first analyze LTENet with different design choices to control self- and cross-reconstruction of LTENet: (1) LTENet (Self) only uses the self-reconstruction of LTENet by setting $\lambda_{\text{self}}=1, \lambda_{\text{cross}}=0$; (2) LTENet (Cross) only uses the cross-reconstruction of LTENet  by setting $\lambda_{\text{self}}=0,\lambda_{\text{cross}}=1$; (3) LTENet (Self \& Cross) uses both self- and cross-reconstruction and removes the mapping loss for the regularization; and (4) LTENet (CD) and LTENet (EMD) are similar to the full model of LTENet in only replacing the CS objective with popular CD and EMD.  All results are summarized in Table \ref{tab:ablation_study}. It is clear that cross-reconstruction contributes most to the final performance and that CS leads to better performance compared to CD and EMD.

\noindent \textbf{Choices of the kernel bandwidth $\sigma$.} As our CS objective is closely relevant to a fixed-bandwidth KDE with Gaussian kernels, it is important to choose a suitable bandwidth fitting the  underlying data distribution of the training dataset --- either too large or too small bandwidth values could lead to degraded performance. As shown in Figure \ref{fig:knn_bandwidth} (right), we chose different bandwidth values by setting $\sigma = 0.01/9, 0.01/3, 0.01, 0.03, 0.09$. The results suggest that $0.01$ is a suitable bandwidth, which we used thereafter for LTENet. It is worth noting that we should adjust the bandwidth accordingly when moving to a new dataset.

\noindent \textbf{Number of nearest neighbors.} Similarly, we can set different numbers of nearest neighbors for the locally linear transformations, i.e., $K=5,10,20,49$. Figure \ref{fig:knn_bandwidth} (left) demonstrates that $K=10$ leads to better performance.  

\begin{table}
\centering
\caption{\textbf{Ablation study on the training sample size}.  We conduct the ablation study under the SURREAL/SHREC setting. }
\label{tab:ablation_study_sizes}
\Huge
\resizebox{0.49\textwidth}{!}{%
\begin{tabular}{lcccccc}
\toprule
Method & $\lambda_{\text{self}}=1$ &  $\lambda_{\text{cross}}=1$ & $\lambda_{\text{reg}}=10$ & size & \textit{acc} $\uparrow$ & \textit{err} $\downarrow \hspace{0.5em}$ \\

\midrule
LTENet (Self) & \cmark &  &  & 2K & 2.3\% & 8.9  \\
LTENet (Cross) &  & \cmark &  & 2K   & 20.6\% & 6.5  \\
LTENet (Self \& Cross) & \cmark & \cmark & & 2K  & 20.6\%  & 5.9 \\
LTENet &  \cmark & \cmark & \cmark  & 2K & 20.7\% & 5.8  \\
\midrule
LTENet (Self) & \cmark &  &  & 230K & 2.2\% & 8.9  \\
LTENet (Cross) &  & \cmark & & 230K   & 19.5\% & 7.2  \\
LTENet (Self \& Cross) & \cmark & \cmark &  & 230K  & 20.5\%  & 6.5 \\
LTENet &  \cmark & \cmark & \cmark  & 230K & 20.9\% & 6.2 \\

\bottomrule
\end{tabular}
}
\end{table}

\begin{table}
\caption{\textbf{Ablation study on the effects of embedding dimension}. We conduct the study under the SURREAL/SHREC setting. }
\label{tab:ablation_study_dimension}
\centering
\begin{tabular}{lcccccc}
\toprule
Method  & Feature dimension & \textit{acc} & \textit{err} $\hspace{0.5em}$ \\

\midrule
LTENet   & 32 & 19.1\% & 6.4  \\
LTENet   & 64  & 19.4\% & 6.2  \\
LTENet  & 128 & 19.7\% & 6.2 \\
LTENet   & 256 & 20.1\% & 6.1  \\
LTENet   & 512 & 20.7\% & 5.8  \\
LTENet   & 1,024 & 20.8\% & 5.9  \\

\bottomrule
\end{tabular}
\end{table}

\noindent \textbf{The impact of training sample size.}  We trained models by increasing the training sample sizes from 2000 to 230,000. Table \ref{tab:ablation_study_sizes} summarizes the experimental results of training models using different training samples. The experimental results show that 2000 samples are sufficient. A subtle difference is found in the slightly increased \textit{err} when training models with $230,000$ training samples, which we suspect is due to more training samples that are symmetric and rotated being included in training, thus creating noisy training signals. Handling symmetry of shapes remains an under-explored research area. For shape correspondence though, there exist some attempts \cite{yoshiyasu2016symmetry,sharma2021learning}, which we believe are promising.

\noindent \textbf{The effect of embedding dimension}. Given $\mathcal{X}$ and $\mathcal{Y}$, we extract their  nonlinear feature embeddings $\mathcal{F}^\mathcal{X},\mathcal{F}^\mathcal{Y} \in \mathbb{R}^{N\times D}$, respectively, via a neural network $\mathcal{F}$. The embedding dimension $D$ should be adjusted to achieve a balance between overfitting and underfitting and efficiency. Table \ref{tab:ablation_study_dimension} demonstrates that the model of $D=512$ gives a good generalization performance while being efficient, which we used thereafter for LTENet.

\section{Discussion}

Our experimental results showed that LTENet achieves superior performance compared to state-of-the-art unsupervised shape correspondence methods. We attribute this to the learning mechanism capable of capturing the underlying structure of the manifold and fully exploiting the local Euclidean geometry of manifolds within local neighborhoods.  Our approach encourages the best locally linear alignment between shape embeddings without ambiguity via the closed-form expression of reconstruction weights. The local linearity used in our approach leads to implicit regularization and universal/canonical embeddings between a pair of shapes in correspondence. We demonstrated the performance gap of our learned embeddings, with and without additional functional map-inspired \textit{globally} linear transformations, and showed consistent improvements made by the additional linear transformations. We observed a mismatch in that embeddings learned from an unsupervised shape correspondence method are not necessarily suitable to be used as basis embeddings in the classic functional map framework.

\subsection{Limitations and Future Work}

\added[id=PAN,comment={}]{
It has been demonstrated that shape correspondence approaches 
are struggling in disambiguating shape symmetries \cite{yoshiyasu2016symmetry,sharma2021learning,lang2021dpc,donati2022deep}. In our work, we also observed the symmetry issue --- it leads to noisy training signals by wrongly matching  components between shapes, e.g., associating the left hand in one human with the right hand in another human due to their opposite orientations. Future work is suggested to handle symmetry by exploiting priors on shapes to impose additional regularization or constraints on embedding learning.}

\added[id=PAN,comment={}]{Many extensions of LLE, such as modified locally linear embedding (MLLE) \cite{zhang2006mlle}, LLE with geodesic distances \cite{varini2005isolle}, and LLE with penalty functions \cite{winlaw2011robust}, have been proposed to further improve LLE. It is promising to incorporate these advanced designs and adapt them to shape correspondence for better performance. The discovered mismatch problem suggests that further exploration of learning embeddings suitable for use as bases in the functional map is a promising direction to establish a unified shape correspondence framework, particularly in the unsupervised learning setting.  
}

\added[id=PAN,comment={}]{In this work, we focus on the matching problem between point cloud shapes. In the future, we propose to extend our method to  matching problems in other modalities, e.g., images and meshes, and cross-modality matching problems, e.g., images to point clouds.
}
\section{Conclusions}

We have presented a novel approach to unsupervised
shape correspondence learning between pairs of point clouds. LTENet is unique in that it introduces an LLE-inspired algorithm that represents maps between these shapes as locally linear transformations in the high-dimensional embedding spaces and leads to the learning of universal/canonical embeddings for shapes in correspondence. The embedding learning is driven by minimizing a suitable divergence measure between the LLE cross-reconstruction of source and target point clouds.  Remarkably, LTENet achieves state-of-the-art performance on standard benchmark datasets while showing strong model generalization across datasets with efficient training and inference. 

\ifCLASSOPTIONcompsoc
  \section*{Acknowledgments}
\else
  \section*{Acknowledgment}
\fi

This work is supported by NSF CNS 1922782.
The opinions, findings, and conclusions expressed in this publication are
those of the authors and not necessarily those of the National Science Foundation.

\ifCLASSOPTIONcaptionsoff
  \newpage
\fi



%


\bibliographystyle{IEEEtran}
\bibliography{ref}

%

\begin{IEEEbiography}[{\includegraphics[width=1in,height=1.25in,clip,keepaspectratio]{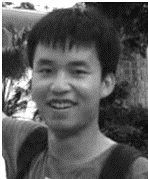}}]%
{Pan He} (Student Member, IEEE) is currently a Ph.D. student in the Department of Computer \& Information Science \& Engineering, University of Florida, Gainesville, FL. His research interests include deep learning and computer vision.
\end{IEEEbiography}
\vspace{-1.65cm}
\begin{IEEEbiography}[{\includegraphics[width=1in,height=1.25in,clip,keepaspectratio]{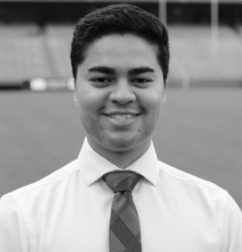}}]%
{Patrick Emami} (Member, IEEE) received the B.Sc.
degree in computer engineering from the University of Florida (UF) in 2016. He received the Ph.D. degree in computer science from UF in 2021.
His research is focused on developing neural algorithms for the analysis and generation of multi-object
spatio-temporal data.
\end{IEEEbiography}
\vspace{-1.65cm}
\begin{IEEEbiography}[{\includegraphics[width=1in,height=1.25in,clip,keepaspectratio]{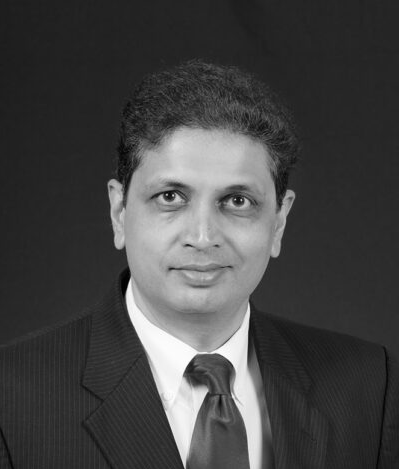}}]%
{Sanjay Ranka} (S’87–M’88–SM’01–F’02) is a Distinguished Professor in the Department of Computer Information Science and Engineering at University of Florida. His current research is on developing algorithms and software using Machine Learning, Internet of Things, GPU Computing, and Cloud Computing for solving applications in Transportation and Health Care. His work has received 14,500+ citations, with an h-index of 61 (Google Scholar). He is a fellow of the IEEE, AAAS, and AIAA (Asia-Pacific Artificial Intelligence Association) and a past member of IFIP Committee on System Modeling and Optimization. He was awarded the 2020 Research Impact Award from IEEE Technical Committee on Cloud Computing. His research is currently funded by NIH, NSF, USDOT, DOE, and FDOT. 
\end{IEEEbiography}
\vspace{-1.4cm}
\begin{IEEEbiography}[{\includegraphics[width=1in,height=1.25in,clip,keepaspectratio]{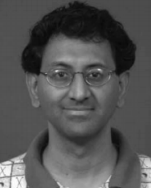}}]%
{Anand Rangarajan} (Member, IEEE) is a Professor in the  Department of Computer and Information Science and Engineering, University of Florida, Gainesville, FL, USA. His research interests are computer vision, machine learning, and the science of consciousness.
\end{IEEEbiography}

\clearpage
\newpage

\appendices


\section{The Constrained Least-Squares Problem} \label{sec:lsq}

Here, we demonstrate that minimizing the cost function in Equation (\ref{eq:lsq}) leads to a closed-form expression of the reconstruction weights $\boldsymbol{W}$. The constraint $\sum_{j=1}^K w_{ij} = 1$ can be written as $\bb{1}^T \bb{w}_i = 1$; therefore, $\bb{x}_i = \bb{x}_i \bb{1}^T \bb{w}_i$. Let the matrix $\mathbb{R}^{D \times K} \ni \bb{X}_i := [\bb{x}_{i1}, \dots, \bb{x}_{iK}]$ denote the $K$ nearest neighbors of $\bb{x}_i$ and we can rewrite the objective in Equation (\ref{eq:lsq}) as
\begin{equation}
    E(\boldsymbol{W}) = \sum_{i=1}^N \Big\Vert \bb{x}_i \bb{1}^T \bb{w}_i - \bb{X}_i \bb{w}_i \Big\Vert_2^2 
\end{equation}

We can simplify the term in $E(\boldsymbol{W})$ as
\begin{equation}
\begin{split}
& \Big\Vert \bb{x}_i \bb{1}^T \bb{w}_i - \bb{X}_i \bb{w}_i \Big\Vert_2^2 \\
&= \Big\Vert (\bb{x}_i \bb{1}^T  - \bb{X}_i) \bb{w}_i \Big\Vert_2^2 \\
&= {\bb{w}_i}^T (\bb{x}_i \bb{1}^T  - \bb{X}_i)^T  (\bb{x}_i \bb{1}^T  - \bb{X}_i) \bb{w}_i \\
&= \bb{w}_i^T \bb{G}_i\, \bb{w}_i,
\end{split}
\end{equation}
where $\bb{G}_i$ denotes the Gram matrix defined as
\begin{equation}\label{eq:gram}
\mathbb{R}^{K \times K} \ni \bb{G}_i := (\bb{x}_i \bb{1}^T  - \bb{X}_i)^T  (\bb{x}_i \bb{1}^T  - \bb{X}_i).
\end{equation}
Therefore,  Equation (\ref{eq:lsq}) is now expressed as
\begin{equation}
\begin{split}\label{eq:lsq_matrix}
    &\minimize_{\{\bb{w}_i\}_{i=1}^N}\quad  \sum_{i=1}^n \bb{w}_i^T \bb{G}_i\, \bb{w}_i, \\
    & \textrm{subject to} \quad  \bb{1}^T \bb{w}_i = 1, \forall i \in \{1, \dots, N\}.
\end{split}    
\end{equation}
The Lagrangian for the objective function in Equation (\ref{eq:lsq_matrix}) is defined as
\begin{equation}
\begin{split}\label{eq:lagrangian}
\mathcal{L} = \sum_{i=1}^N \bb{w}_i^T \bb{G}_i\, \bb{w}_i - \sum_{i=1}^N \lambda_i\, (\bb{1}^T \bb{w}_i - 1).
\end{split}    
\end{equation}
We then calculate the gradients  $\frac{\partial \mathcal{L}}{\partial \bb{w}_i} \in \mathbb{R}^{k}$ and $ \frac{\partial \mathcal{L}}{\partial \lambda_i} \in \mathbb{R}$ and set them to zero, which gives
\begin{align}
 & \frac{\partial \mathcal{L}}{\partial \bb{w}_i} = 2 \bb{G}_i \bb{w}_i - \lambda_i \bb{1} \overset{\text{set}}{=} \bb{0}, \nonumber\\
&\implies \bb{w}_i = \frac{1}{2} \bb{G}_i^{-1} \lambda_i \bb{1} = \frac{\lambda_i}{2} \bb{G}_i^{-1} \bb{1}. \label{eq:grad_w1} \\
&\implies \bb{1}^T \bb{w}_i  = \frac{\lambda_i}{2} \bb{1}^T \bb{G}_i^{-1} \bb{1}. \label{eq:grad_w2} \\
& \frac{\partial \mathcal{L}}{\partial \lambda} = \bb{1}^T \bb{w}_i - 1 \overset{\text{set}}{=} 0 \nonumber \\
& \implies \bb{1}^T \bb{w}_i = 1.  \label{eq:grad_lambda}
\end{align}    
Using the transitive equality of Equation (\ref{eq:grad_lambda}) and Equation (\ref{eq:grad_w2}), we obtain
\begin{align}\label{eq:lambda}
\frac{\lambda_i}{2} \bb{1}^T \bb{G}_i^{-1} \bb{1} = 1 \implies \lambda_i = \frac{2}{\bb{1}^T \bb{G}_i^{-1} \bb{1}}. 
\end{align}
Substituting Equation (\ref{eq:lambda}) into Equation  (\ref{eq:grad_w1}) gives
\begin{align}\label{eq:lsq_solution}
\bb{w}_i = \frac{\lambda_i}{2} \bb{G}_i^{-1} \bb{1} = \frac{\bb{G}_i^{-1} \bb{1}}{\bb{1}^T \bb{G}_i^{-1} \bb{1}}.
\end{align}

In practice, a $\ell_2$ norm-based regularization term is usually added to Equation (\ref{eq:lsq_matrix}), which gives
\begin{equation}
\begin{split}
    &\minimize_{\{\bb{w}_i\}_{i=1}^N}\quad  \sum_{i=1}^N \bb{w}_i^T \bb{G}_i\, \bb{w}_i + \gamma \|\bb{w}_i\|_2^2, \\
    & \textrm{subject to} \quad  \bb{1}^T \bb{w}_i = 1, \forall i \in \{1, \dots, N\}.
\end{split}    
\end{equation}
Following a similar derivation, the optimal weights can be obtained as
\begin{align}\label{eq:lsq_solution_l2}
\bb{w}_i = \frac{(\bb{G}_i+ \gamma \bb{I})^{-1} \bb{1}}{\bb{1}^T (\bb{G}_i+ \gamma \bb{I})^{-1} \bb{1}},
\end{align}
which could better handle noise. In fact, this regularization leads to a numerically stable solution by avoiding the possible singularity of $\bb{G}_i$ \cite{winlaw2011robust,ghojogh2020locally}. Empirically, we set $\gamma=1$. 

Following the above steps, we can similarly derive the solution of Equation (\ref{eq:cross_reverse_weight}) in the main paper.

\section{The Closed-form Expression for the CS divergence} \label{sec:cs}

Inspired by the CS inequality, the CS divergence measure \cite{jenssen2005optimizing} is defined as 
\begin{equation}
\begin{split}
    \mathcal{D}_{CS}(q,p)  & = - \log \Big( \frac{\int q(x) p(x) dx}{\sqrt{\int q(x)^2 dx \int p(x)^2 dx}} \Big)  \\
  & = - \log  \int  q(x) p(x) dx +  0.5 \log  \int q(x)^2 dx \\ &  \quad   +  0.5 \log \int p(x)^2 dx, 
\end{split}
\end{equation}
which is symmetric for any two PDFs $q$ and $p$ such that $0 \le \mathcal{D}_{CS} < \infty$ where the minimum is obtained iff $q(x) = p(x)$.

Given $p_\mathcal{X}(x) = \frac{1}{N} \sum_{i=1}^N \mathsf{K} (\frac{\bb{x}-\bb{x}_i}{\sigma})$ and $p_\mathcal{Y}(x) = \frac{1}{N} \sum_{j=1}^N \mathsf{K} (\frac{\bb{x}-\bb{y}_j}{\sigma})$ where the Gaussian kernel $G_\sigma (\bb{x}, \bb{y})= \frac{1}{\sqrt{2 \pi} \sigma} \exp{(-\frac{\lVert \bb{x} - \bb{y}\rVert}{2 \sigma^2})}$ is used as the kernel function $\mathsf{K}$. According to the convolution theorem for Gaussian functions, it has been shown in \cite{bromiley2003products} that the following relation holds:

\begin{equation}
    \int  G_\sigma (\bb{x}, \bb{x}_i) G_\sigma (\bb{x}, \bb{y}_j) dx = G_{\sqrt{2}\sigma} (\bb{y}_j, \bb{x}_i).
\end{equation}

We then substitute Gaussian kernel PDF estimators into $q(x), p(x)$ and obtain
\begin{equation}
 \begin{split}
    \int  q(x) p(x) dx  &  \approx \int  p_\mathcal{X}(x) p_\mathcal{Y}(x) dx \\
    & = \int \frac{1}{N} \sum_{i=1}^N \mathsf{K} (\frac{\bb{x}-\bb{x}_i}{\sigma}) \sum_{j=1}^N  \frac{1}{N} \mathsf{K} (\frac{\bb{x}-\bb{y}_i}{\sigma}) dx \\
    & = \frac{1}{N^2}  \sum_{j=1}^N  \sum_{i=1}^N \int  G_\sigma (\bb{x}, \bb{x}_i) G_\sigma (\bb{x}, \bb{y}_j) dx \\
    & = \frac{1}{N^2} \sum_{j=1}^N \sum_{i=1}^N   G_{\sqrt{2}\sigma} (\bb{y}_j, \bb{x}_i).
\end{split}   
\end{equation}

Now, we perform a similar calculation to $\int q(x)^2 dx$ and $\int p(x)^2 dx$ and obtain
\begin{equation}
 \begin{split}
    \int q(x)^2 dx = \frac{1}{N^2}  \sum_{i'=1}^N \sum_{i=1}^N  G_{\sqrt{2}\sigma} (\bb{x}_{i'},\bb{x}_i), \\ 
    \int p(x)^2 dx = \frac{1}{N^2} \sum_{j'=1}^N \sum_{j=1}^N  G_{\sqrt{2}\sigma}. (\bb{y}_j', \bb{y}_j).
\end{split}   
\end{equation}

Based on them, we can express the  Cauchy-Schwarz  divergence as 
\begin{equation}\label{eq:cs_appendix}
\small
\begin{split}
\mathcal{D}_{CS}(p_\mathcal{X},p_\mathcal{Y})   = & - \log \sum_{j=1}^N \sum_{i=1}^N   G_{\sqrt{2}\sigma} (\bb{y}_j, \bb{x}_i) \\
& + 0.5 \log \sum_{i'=1}^N \sum_{i=1}^N  G_{\sqrt{2}\sigma} (\bb{x}_{i'},\bb{x}_i) \\ 
  & + 0.5 \log \sum_{j'=1}^N \sum_{j=1}^N  G_{\sqrt{2}\sigma}. (\bb{y}_j', \bb{y}_j) \\
\end{split}   
\end{equation}

$\mathcal{D}_{CS}(p_\mathcal{\hat{Y}}, p_\mathcal{Y})$ in Equation (\ref{eq:cs_divergence}) can be similarly expressed following $\mathcal{D}_{CS}(p_\mathcal{X},p_\mathcal{Y})$ in Equation (\ref{eq:cs_appendix}).

\section{The Closed-form Expression for the Optimal Linear Transformation}\label{sec:optimal}

\added[id=PAN,comment={}]{
Given two fixed embeddings $\mathcal{F}^\mathcal{X}, \mathcal{F}^\mathcal{Y}$, we want to find 
a linear transformation $\mathcal{A}_{\mathcal{XY}}$ for a fixed point-to-point correspondence matrix $\Pi_\mathcal{XY}$, which satisfies
\begin{equation}
\begin{split}
     \mathcal{A}_{\mathcal{XY}} (\mathcal{F}^\mathcal{X})^T & = (\Pi_\mathcal{XY} \mathcal{F}^\mathcal{Y})^T  \iff \\
         \mathcal{F}^\mathcal{X}  \mathcal{A}_{\mathcal{XY}}^T & = \Pi_\mathcal{XY} \mathcal{F}^\mathcal{Y}. 
\end{split}   
\end{equation}
This can be formulated as a least squares problem defined as 
\begin{equation}
\mathcal{A}_{\mathcal{XY}}^T = \argmin_{A^T} \|\mathcal{F}^\mathcal{X}   A^T - \Pi_\mathcal{XY} \mathcal{F}^\mathcal{Y} \|_2,
\end{equation}
which gives a close-formed solution
\begin{equation}
   \mathcal{A}_{\mathcal{XY}} = A = ((\mathcal{F}^\mathcal{X})^\dagger \Pi_\mathcal{XY} \mathcal{F}^\mathcal{Y})^T.
\end{equation}
where $\dagger$ denotes the Moore Penrose pseudo-inverse.
}

\begin{figure*}
\begin{center}
\begin{tabular}{cccc}
\multicolumn{4}{c}{\includegraphics[width=\textwidth]{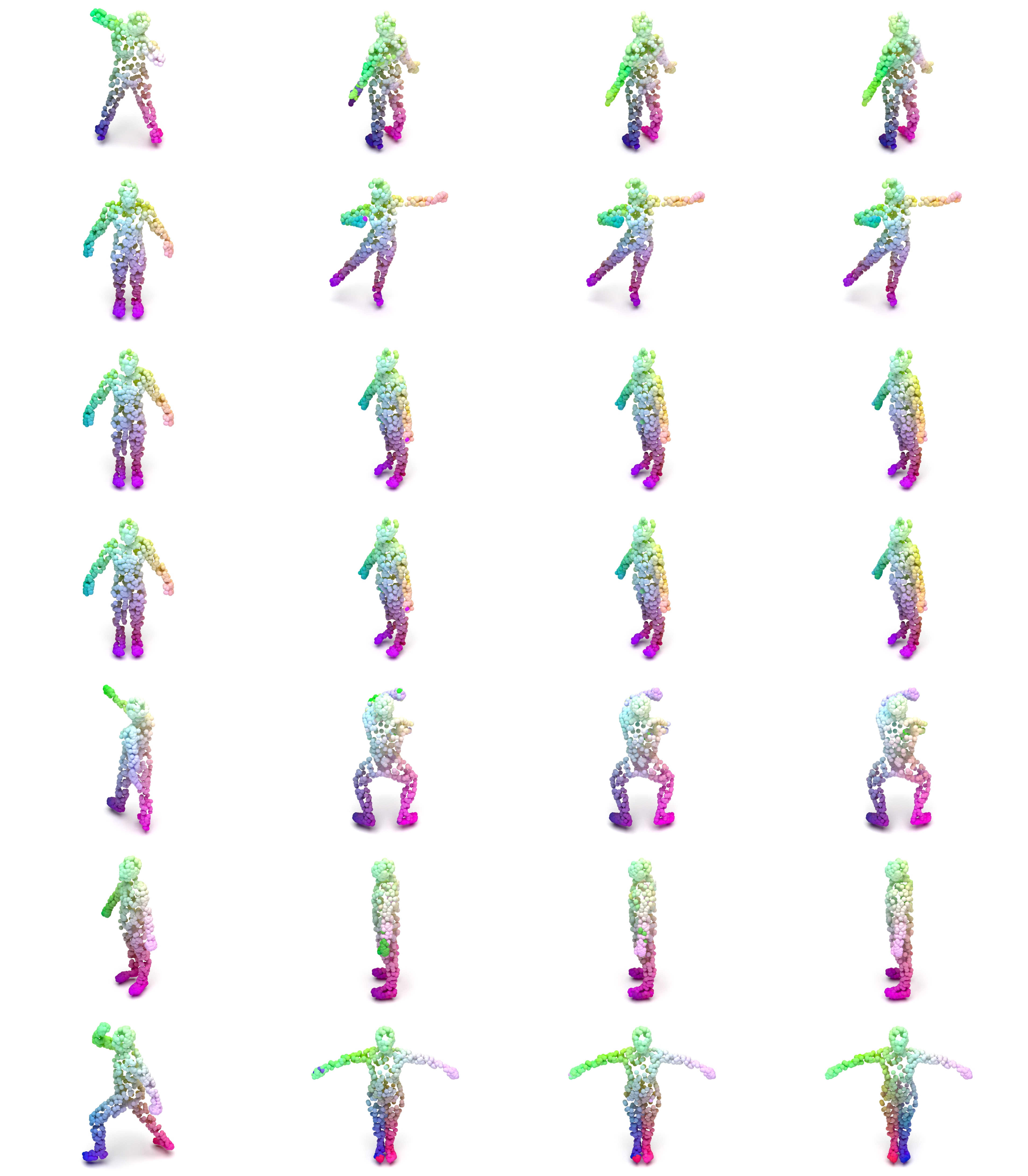}} \\
\whitetext{aaaaaaaa}Reference target & \whitetext{aaaaaaaaaaaa}DPC \cite{lang2021dpc} & \whitetext{aaaaaaaaaaaaa} LTENet (ours) & \whitetext{aaaaaaa} Ground-truth \\
\end{tabular}
\end{center}
\caption{\textbf{Visual examples of SHREC
test pairs.} The experiment follows the SURREAL/SHREC setting. The shape correspondence mappings are color-coded. In the first six rows, we compare our LTENet against the state-of-the-art DPC \cite{lang2021dpc} and show the clear improvement made by our LTENet on generating more accurate correspondence predictions. The last row shows the typical failure example of LTENet and DPC \cite{lang2021dpc}. Handling symmetry and rotation of shapes remains a challenging problem and requires future investigation. }\label{fig:human_more}
\end{figure*}

\begin{figure*}
  \begin{minipage}{0.33\linewidth} 
    \centering
    \includegraphics[width=\textwidth]{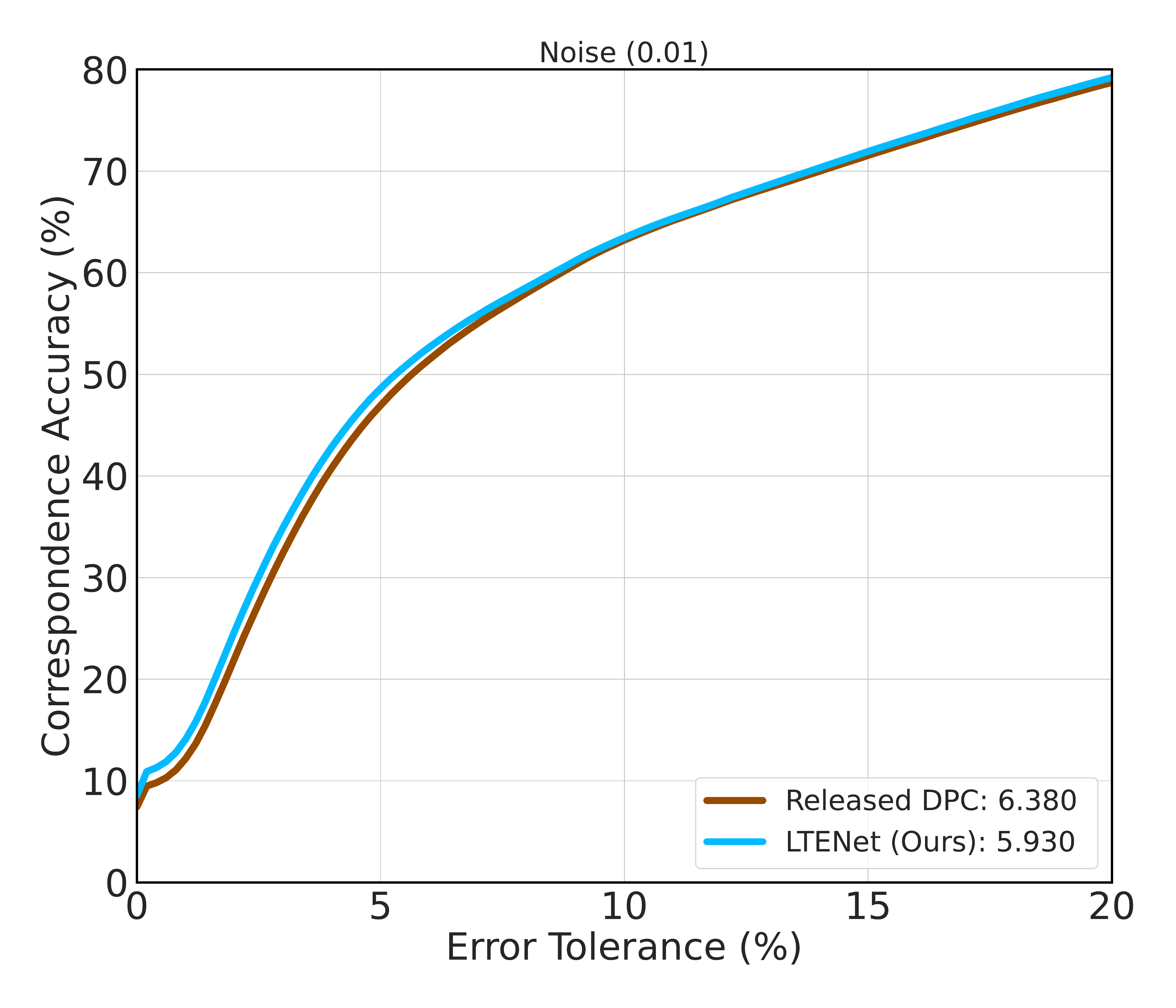}
    (a)
  \end{minipage}
 \hfill
  \begin{minipage}{0.33\linewidth}
   \centering
    \includegraphics[width=\textwidth]{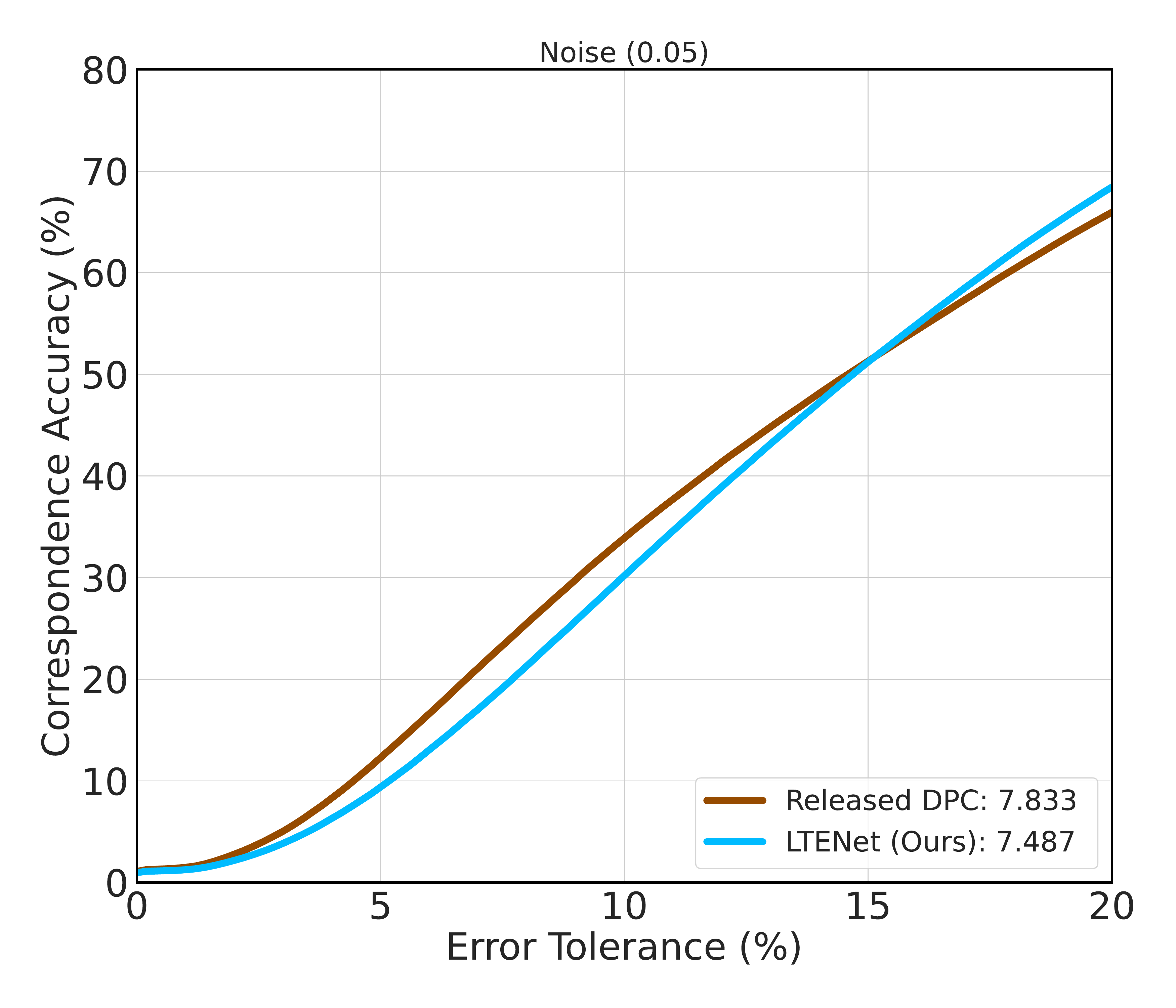}
    (b)
  \end{minipage}
  \hfill
  \begin{minipage}{0.33\linewidth}
   \centering
    \includegraphics[width=\textwidth]{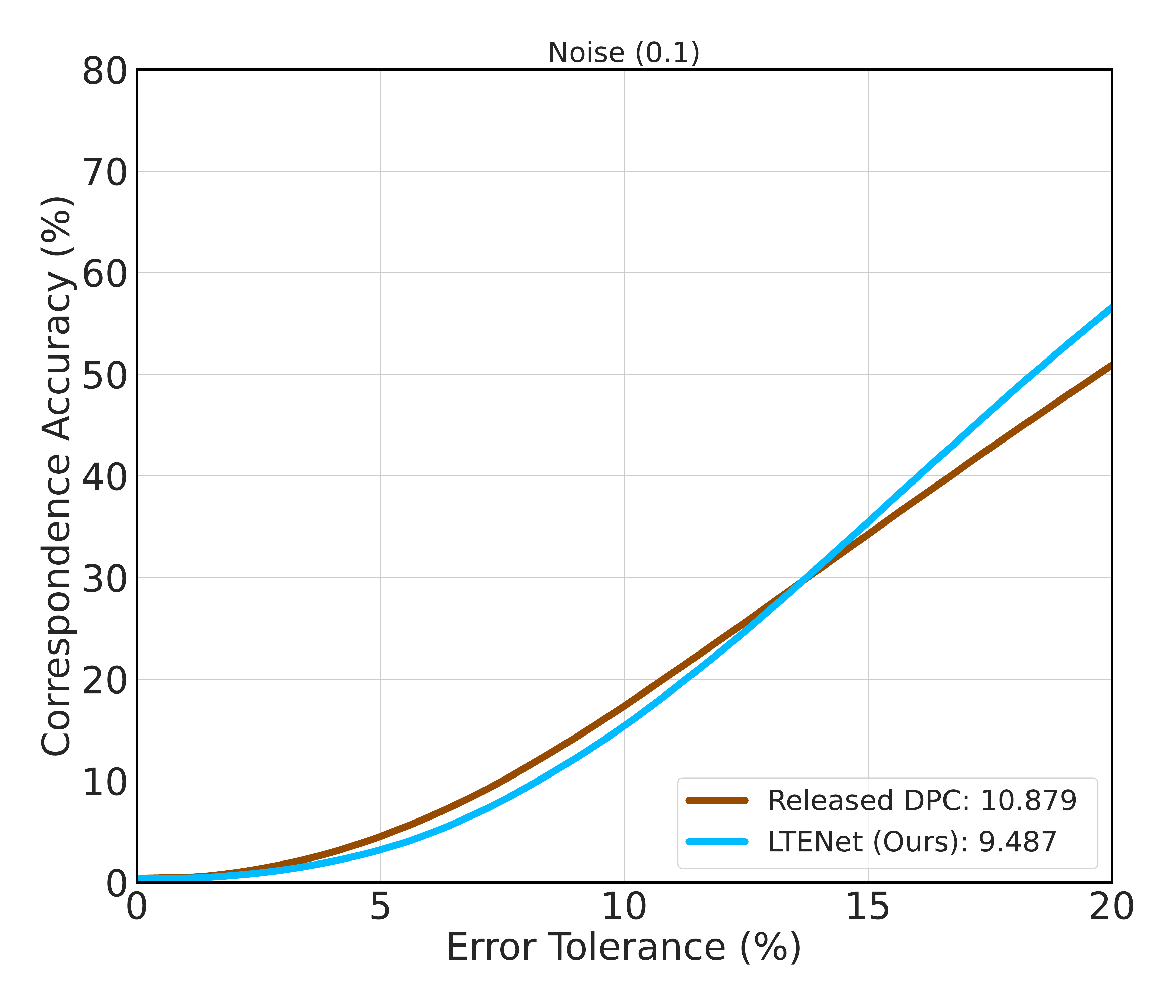}
    (c)
  \end{minipage}
  \caption{The evaluation of correspondence prediction of SHREC test point clouds in the SURREAL/SHREC setting with additional noise. From (a) to (c), we gradually add stronger Gaussian noises with zero means and larger standard deviations, i.e., $0.01, 0.05, 0.1$, to source shapes.}\label{fig:noise_shrec}
\end{figure*}

\begin{figure*}[tb!]
\begin{center}
\begin{tabular}{cccc}
\multicolumn{4}{c}{\includegraphics[width=\textwidth]{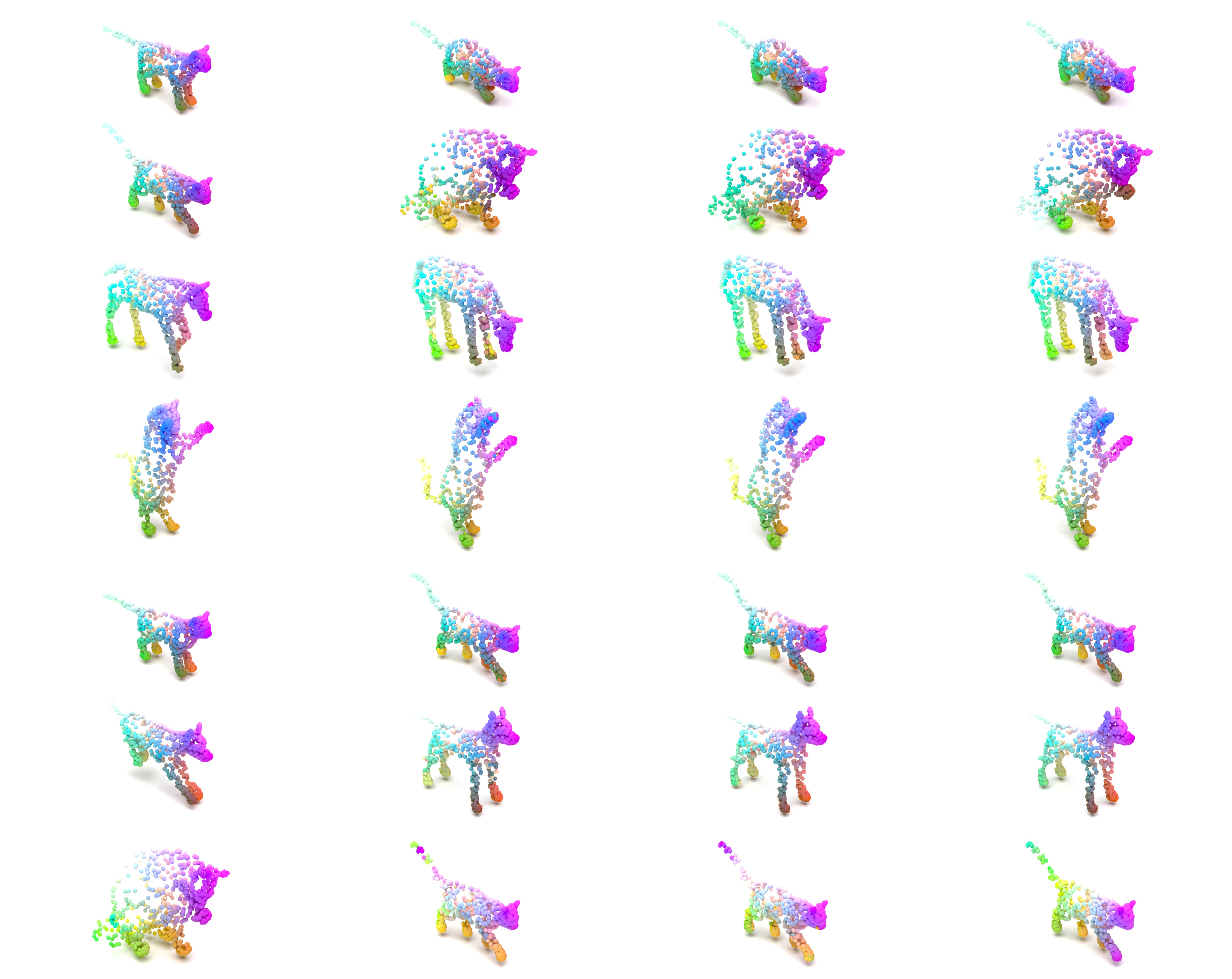}} \\
\whitetext{aaaaaaaa}Reference target & \whitetext{aaaaaaaaaaaaa}DPC \cite{lang2021dpc} & \whitetext{aaaaaaaaaaaaaaa} LTENet (ours) & \whitetext{aaa} Ground-truth \\
\end{tabular}
\end{center}
\caption{\textbf{Visual examples of
TOSCA test pairs.} The experiment follows the SMAL/TOSCA setting. The shape correspondence mappings are color-coded. The proposed LTENet generates more accurate correspondence predictions  compared to DPC. }\label{fig:non_human_more}
\end{figure*}




\end{document}